\documentclass{article}

\pdfoutput=1

\usepackage{neurips_2023}

\usepackage[utf8]{inputenc} 
\usepackage[T1]{fontenc}    
\usepackage{url}            
\usepackage{amsfonts}       
\usepackage{nicefrac}       

\usepackage{amsmath}
\usepackage{amssymb}
\usepackage{mathtools}
\usepackage{amsthm}
\usepackage{microtype}
\usepackage{graphicx}
\usepackage{subcaption}
\usepackage{booktabs} 
\usepackage{bm, bbm}
\usepackage{xcolor}
\usepackage{wrapfig}
\usepackage{subfiles}
\usepackage{thm-restate}
\usepackage{multirow}
\usepackage{makecell}
\usepackage{algorithm}
\usepackage{algpseudocode}
\usepackage{graphics}
\usepackage{caption}
\usepackage[numbers]{natbib}

\theoremstyle{plain}
\newtheorem{lemma}{Lemma}

\newtheorem{assumption}{Assumption}

\newtheorem{proposition}{Proposition}

\title{Bandit Learning to Rank with Position-Based Click Models: Personalized and Equal Treatments}

\author{%
  Tianchen Zhou$^1$ \hspace{2pt}Jia Liu$^1$ \hspace{2pt}Yang Jiao$^2$ \hspace{2pt}Chaosheng Dong$^2$ \hspace{2pt}Yetian Chen$^2$ \hspace{2pt}Yan Gao$^2$ \hspace{2pt}Yi Sun$^2$\\
  \\
  $^1$Dept. of Electrical and Computer Engineering, The Ohio State University, OH, USA; \\
  $^2$Amazon, WA, USA\\
  \texttt{zhou.2220@osu.edu; liu@ece.osu.edu;}\\
  \texttt{$\lbrace$jaoyan; chaosd; yetichen; yanngao; yisun$\rbrace$@amazon.com}
}

\begin{document}

\maketitle

\begin{abstract}
Online learning to rank (ONL2R) is a foundational problem for many recommender systems and has received increasing attention in recent years. 
Among the existing approaches for ONL2R, a natural modeling architecture is the multi-armed bandit (MAB) framework coupled with the position-based click model. 
However, developing efficient online learning policies for MAB-based ONL2R with position-based click models is highly challenging due to the combinatorial nature of the problem, partial observability in the position-based click model, and the complex coupling between the ranking position preferences and the mean reward of the ranking recommendations.
To date, results in MAB-based ONL2R with position-based click models remain rather limited, which motivates us to fill this gap in this work. 
Our main contributions in this work are threefold: i) We propose the first general MAB framework that captures all key ingredients of ONL2R with position-based click models. 
Our model considers personalized and equal treatments in ONL2R ranking recommendations, both of which are widely used in practice; 
ii) Based on the above analytical framework, we develop two unified greed- and upper-confidence-bound (UCB)-based policies called GreedyRank and UCBRank, each of which can be applied to personalized and equal ranking treatments; 
and iii) We show that both GreedyRank and UCBRank enjoy $\mathcal{O}(\sqrt{t}\ln t)$ and $\mathcal{O}(\sqrt{t\ln t})$ {\em anytime} sublinear regret for personalized and equal treatment, respectively. 
For the fundamentally hard equal ranking treatment, we identify classes of collective utility functions and their associated sufficient conditions under which $\mathcal{O}(\sqrt{t}\ln t)$ and $\mathcal{O}(\sqrt{t\ln t})$ anytime sublinear regrets are still achievable for GreedyRank and UCBRank, respectively. 
Our numerical experiments also verify our theoretical results and demonstrate the efficiency of GreedyRank and UCBRank in seeking the optimal action under various problem settings.

\end{abstract}


\section{Introduction} \label{sec:introduction}

Learning to rank (L2R) is a foundational problem for recommender systems \cite{sorokina2016amazon}.
Solving an L2R problem amounts to understanding and predicting users’ browsing and clicking behaviors, so that the system can accordingly provide an optimal ranking of items to recommend to users with the aim to maximize certain rewards or utilities for the system.
In the literature, L2R has been relatively well studied in the offline supervised setting, where a dataset is used to train a model in an offline fashion and then the learned model is used for ranking prediction.
However, offline L2R can only provide static results that cannot adapt to real-time data and temporal changes of the underlying ground truth.
Therefore, in recent years, online L2R (ONL2R) has received increasing attention~\cite{agichtein2006improving}.
Among the existing approaches for ONL2R, the multi-armed bandit (MAB) framework is one of the most popular since it closely models the sequential interactions between the recommender system and the users (e.g., \cite{radlinski2008learning,lagree2016multiple,zoghi2017online,lattimore2018toprank}).
In MAB-based ONL2R methods, the goal of the learner is to understand the click models of a spectrum of different user types through {\em bandit} feedback (i.e., data are collected in real-time through action-reward interactions rather than preexisting)~\cite{chuklin2015click}. 
Based on the bandit feedback, the system follows an online learning policy and {\em iteratively} adjusts the ranking of items to the next arriving user to maximize its long-term accumulative reward.

Clearly, a key component in MAB-based ONL2R is the click model. A natural choice of click model is the so-called {\em position-based click model} \cite{richardson2007predicting, lagree2016multiple}, where each ranking position is associated with a preference probability of being observed and clicked.
Studies have shown that user actions are highly influenced by webpage layouts or ranking positions: if a listing is not displayed in some particular area of the web layout, then the odds of being seen by a searcher are dramatically reduced \cite{hotchkiss2005eye}.
Position-based click model is also shown to be closely related to various popular ranking quality metrics for recommender systems, such as normalized discounted cumulative gain (NDCG)\cite{valizadegan2009learning, wang2013theoretical}.

However, developing efficient online learning policies for MAB-based ONL2R with position-based click models is highly non-trivial due to the following technical challenges:
First, MAB-based ONL2R problems with position-based click models are {\em combinatorial} in nature, which means that their offline counterparts are already NP-hard in general.
Second, due to the  multi-user nature, ranking recommendations for MAB-based ONL2R problems are complicated by the philosophical debate whether we should provide personalized or equal treatments to different users, both of which are common in practice.
To date, there remains a lack of rigorous understanding on how different types of ranking treatments could affect MAB-based ONL2R policy design.
Third, unlike conventional MAB problems, there is a fundamental {\em partial observability} challenge in MAB-based ONL2R policy design.
Specifically, in many real-world recommender systems, if a user does not click on any displayed ranked item, the system will not receive any feedback on which ranked position has been observed by the user.
This uncertainty creates an extra layer of challenge in MAB-based ONL2R.
Last but not least, in MAB-based ONL2R with position-based click models, there is a complex coupling between each ranking position's observation preference and mean reward of each arm, both of which are not only unknown and need to be learned, but also heterogeneous across user types.
Due to these challenges, results for MAB-based ONL2R with position-based click models are rather limited in the literature (see Section~\ref{sec:related} for more detailed discussion), which motivates us to fill this gap in this work.

The main contribution of this paper is that we propose a series of new MAB policy designs, which overcome the aforementioned challenges.
The key results of this paper are summarized as follows:

\begin{list}{\labelitemi}{\leftmargin=0.5em \itemindent=-0.0em \itemsep=.1em}
\vspace{-.1in}

\item We propose the first general MAB framework that captures all key ingredients of ONL2R with position-based click models: i) two regret notions that characterize personalized and equal treatments in ranking recommendations; ii) the coupling between position preferences and mean arm rewards; and iii) partial observability of ranking position preferences.
This general framework enables our rigorous policy design and analysis for MAB-based ONL2R with position-based click models.

\item Based on the above general MAB-based ONL2R framework with position-based click models, we develop two {\em unified} greedy- and upper-confidence-bound (UCB)-based policies, each of which works for personalized and equal ranking treatments.
For personalized treatment for ranking recommendations, we show that our greedy- and UCB-based policies achieve $\mathcal{O}(\sqrt{t}\ln t)$ and $\mathcal{O}(\sqrt{t\ln t})$ {\em anytime} sublinear regrets, respectively.

\item We show that the MAB policy design for the equal treatment case is more challenging, which may require solving an NP-hard problem in each time step depending on the collective utility function for social welfare.
To address this challenge, we identify classes of utility functions and establish their associated sufficient conditions of approximation accuracy, under which $\mathcal{O}(\sqrt{t}\ln t)$ and $\mathcal{O}(\sqrt{t\ln t})$ {\em anytime} sublinear regrets are still achievable for greedy- and UCB-based policies, respectively.

\end{list}

\section{Related Work} \label{sec:related}

We provide an overview on three closely related fields to put our work into comparative perspectives.

\textbf{1) MAB-Based ONL2R:} 
As mentioned earlier, research on MAB-based ONL2R remains in its infancy.
To our knowledge, the first work that studied MAB-based ONL2R was reported in \cite{radlinski2008learning}, which, however, is based on the cascade click models.
Later, MAB-based ONL2R with position-based click model was considered in \cite{lagree2016multiple}, which established an $\Omega(\log T)$ regret lower bound   for their model and proposed algorithms with a matching regret upper bound.
Although sharing some similarity to ours, the position-based click model in \cite{lagree2016multiple} is a simpler model setting, which assumes {\em known} position preference.
In contrast, the unknown position preferences in our work requires extra learning besides conventional arm mean estimation, causing non-trivial policy design and performance trade-off.
Generalized click model encompassing both position-based and cascade click models was proposed in \cite{zoghi2017online}, where the authors also developed a BatchRank policy with a gap-dependent upper bound on the $T$-step regret.
BatchRank was later outperformed by the TopRank policy proposed in \cite{lattimore2018toprank} in both cascade and position-based click models.
We note that all these existing works make a strong and unrealistic assumption that user behavior is {\em homogeneous}, and they all aim to optimize a standard MAB objective, i.e., maximizing total clicks.
In contrast, we consider a more general setting with multiple user types, and design policies with two practical popular objectives: personalized treatment and equal treatment. 

\textbf{2) Combinatorial Semi-Bandit:} Similar to our model, combinatorial semi-bandit (CSB) also considers multiple arms being pulled with semi-bandit feedback at each round \cite{kveton2015combinatorial, chen2016combinatoriala, chen2016combinatorialb, wang2018thompson}.
For example, the work in \cite{chen2016combinatorialb} studied the general CSB framework, and proposed a UCB-style policy CUCB with regret upper bound.
Later, the work in \cite{wang2018thompson} developed a policy based on Thompson sampling.
Also, the work in \cite{kveton2015combinatorial} derived two upper bounds on the $n$-step regret of policy CombUCB1, while proving a matching lower bound using a partition matroid bandit.
We note that the CSB setting differs from ours in two key aspects: i) while CSB considers a subset of arms as a super arm at each round, our setting additionally considers the ranking {\em within} the super arm that also affects the reward; ii) the reward in our setting is based on {\em two unknown parameters}: position preferences and arm means, which cannot be directly estimated separately due to partial observation of the user feedback.
These complications are unseen in conventional CSB settings.

\textbf{3) Personalized and fairness-aware MAB:} 
Personalization is an important topic in online recommender systems.
Recent work in \cite{ermis2020learning} formulated this problem as a contextual bandit for online ranking.
Other works in personalized MAB include carousel personalization in music streaming apps~\cite{bendada2020carousel} and personalized educational content recommendation to maximize learning gains~\cite{segal2018combining}.
Among these related works on personalized MAB, our work is the first to propose policies with performance guarantee on position-based online ranking.
The equal treatment setting in our work is also related to fairness-aware MAB.
In the fairness-aware MAB literature, most works focus on reward maximization with fairness being modeled as constraints.
For example, individual fairness constraints were imposed in the MAB model in \cite{joseph2016fairness}. 
Some other models considered player-side group fairness constraints (e.g., \cite{schumann2019group,schumann2022group,huang2022achieving}).
Another line of research integrates the learning objectives with fairness considerations.
For example, multi-agent bandits with Nash social welfare as the objective function were considered in \cite{hossain2021fair}.
Although not interpreted from a fairness perspective, bandit convex optimization problems with concave objective functions were considered in \cite{hazan2014bandit, bubeck2016multi, bubeck2017kernel}.
In contrast, our policies learn rankings that maximize social welfare in terms of generic fairness utilities that are not necessarily concave and only assumed to be Lipschitz continuous. 


\section{System model and problem formulation}
\label{model}

\begin{wrapfigure}{r}{0.55\textwidth}
\vspace{-.03in}
\begin{center}
  \includegraphics[trim=0 0 0 35, width=.55\textwidth]{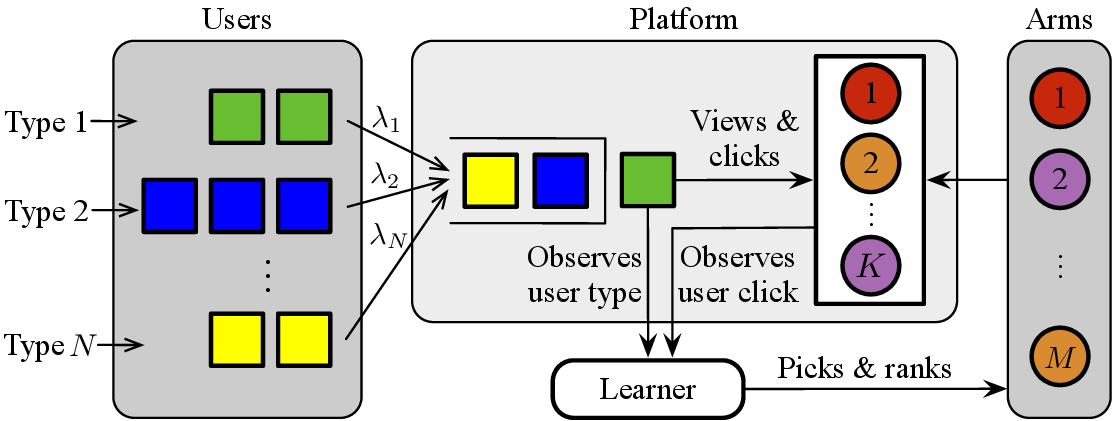}
\end{center}
\caption{System model of MAB-based ONL2R with position-based click models.}
\label{fig:model}
\vspace{-.2in}
\end{wrapfigure}

\textbf{1) System Setup:}
As shown in Fig.~\ref{fig:model}, consider a stochastic bandit setting with a set of user types $[N]\coloneqq \lbrace 1,\ldots, N\rbrace$, a set of arms $[M]\coloneqq \lbrace 1,\ldots, M\rbrace$, and a set of ranking positions $[K]\coloneqq \lbrace 1,\ldots, K\rbrace$, where $K\leq M$.
Each user type $i$ has an arrival rate $\lambda_i>0$.
Without loss of generality, the arrival rates are normalized such that $\sum_{i\in[N]}\lambda_i=1$.
For each position $k \in [K]$, each user type $i \in [N]$ has a position preference $\rho_{i,k}\in[0,1]$, which represents the chance that user type $i$ observes position $k$.
We note that such position preferences have been widely observed in practice.
For example, demographic studies \cite{hotchkiss2005eye} showed that there exist many different user's position-based action patterns that are related to user's gender, education, age, etc.
Such user-specific position-based patterns include ``quick click,'' ``the linear scan,'' ``the deliberate scan,'' ``the pick up search,'' etc.
Thus, the same position can have different preference rates over different groups of people.
For each arm $j \in [M]$, each user type $i$ has a Bernoulli reward distribution $D_{i,j}$ with mean $\mu_{i,j}\in[0,1]$, where $\mu_{i,j}$ can be interpreted as the click rate of arm $j$ if observed by user type $i$.
We assume that the arrival rates $\lambda_i$, the position preferences $\rho_{i,k}$, and the arm means $\mu_{i,j}$ are all {\em unknown} to the learner.
With the basic system setup, we are now in a position to describe the unique key features of our MAB model.

\smallskip
{\bf 2) Agent-User Interaction Protocol:}
At time step $t$, a user of type $I(t)=i$ arrives with probability $\lambda_i$.
With a slight abuse of notation, we also use $I(t)$ to denote the current user at time step $t$.
The learning agent observes $I(t)$, then picks a $K$-sized subset of arms from $[M]$ and determines a $K$-permutation $\sigma_t\in P^M_K$, where $\sigma_t(j)$ represents the ranked position of arm $j$.
Next, user $I(t)$ randomly observes an arm $J(t)$ at ranked position $\sigma_t(J(t))$ with probability $\rho_{I(t),\sigma_t(J(t))}$, and then clicks arm $J(t)$ with probability $\mu_{I(t), J(t)}$.
The learner receives a reward $r(t) = 1$ if some position is clicked, and $r(t)=0$ otherwise.
Clearly, we have $\sum_{k\in[K]} \rho_{i,k} = 1$.
We note that if user $I(t)$ chooses not to click any arm $J(t)$, then the learner receives {\em no information} regarding which position has been observed.
In other words, a reward of $r(t)=0$ can happen by the fact that any random arm in $\sigma_t$ is observed but not clicked. 
This partial observation setting closely follows the reality in most recommender systems, while it also makes the estimation of arm means and position preferences much harder.
We illustrate this learner-user interaction in Fig.~\ref{fig:model}.
A policy $\pi$ sequentially makes decisions on the permutation $\sigma_t$, and observes stochastic rewards $r(t)$ over time $t$.

\smallskip
{\bf 3) Regret Modeling:}
In this paper, we consider two MAB-based ONL2R problems with two different ranking recommendation settings: personalized and equal treatments.

{\em 3-a) Personalized Treatment:} In this setting, the learning policy makes ranking personalized decisions according to the arrived user types.
Thus, an optimal policy always recommends the optimal permutation denoted by $\sigma_i^*$ regarding the arrived user type $i$ to achieve {\em maximum user satisfaction}.
We note, however, that personalized treatment may not be fair to the arms since some arms may never be shown to any user type.
At time $t$, an expected regret $\mathbb{E}[R(t)]$ incurred by a policy $\lbrace \sigma_t \rbrace_t$ is defined as follows:
\begin{equation*}
	\mathbb{E}[R(t)] = \sum_{s=1}^t \left( \langle \bm{\rho}_{I(s)}, \bm{\mu}_{I(s),\sigma^*} \rangle - \mathbb{E}[\langle \bm{\rho}_{I(s)}, \bm{\mu}_{I(s),\sigma_s} \rangle] \right),
\end{equation*}
where we use the notation of vector $\bm{\rho}_i=[\rho_{i,1}, \rho_{i,2}, \ldots, \rho_{i,K}]^{\top}$, and use the notation $\bm{\mu}_{i,\sigma} = [\mu_{i,\sigma^{-1}(1)}, \ldots, \mu_{i,\sigma^{-1}(K)}]^{\top}$ to denote the vector of arm means ranked by permutation $\sigma$, where $\sigma^{-1}: [K] \rightarrow [M]$ is the reverse mapping of the permutation $\sigma$.


{\em 3-b) Equal treatment:} Motivated by fairness, an equal treatment ranking policy makes an identical ranking decision among all user groups to avoid discrimination over sensitive groups.
This is because, in some scenarios, service providers are legally required to treat all sensitive groups the same way by recommending the same ranking content.
In the equal treatment setting, we measure its ranking quality by a general collective utility function (CUF) defined as follows:
\begin{equation*}
	\Gamma(\sigma) \triangleq \sum_{i\in[N]} \lambda_i \cdot f\bigg(\sum_{j\in \mathrm{M}_{\sigma}} \rho_{i,\sigma(j)}\cdot \mu_{i,j}\bigg),
\end{equation*}
where $\mathrm{M}_{\sigma}$ is the set of arm indices in permutation $\sigma$, and $f(\cdot)$ is a generic utility function that transforms the CUF to different social welfare criteria, which we will discuss later.
An optimal policy always recommends a universal optimal permutation denoted by $\sigma^*$ that maximizes CUF, i.e., $\sigma^* = \arg\max_{\sigma\in P^M_K} \Gamma(\sigma)$.
We note that under this setting, due to multiple user types and their unequal arm means, the optimal permutation $\sigma^*$ may not be a decreasingly ordered arm list.
At time $t$, an expected regret $\mathbb{E}[R(t)]$ incurred by a policy $\lbrace \sigma_t \rbrace_t$ is defined as follows:
\begin{equation*}
	\mathbb{E}[R(t)] = t\cdot\Gamma(\sigma^*) - \sum_{s=1}^t \mathbb{E}\left[\Gamma(\sigma_s)\right].
\end{equation*}

Here, we provide two common examples of CUF: i) the utilitarian CUF and ii) the Nash CUF~\cite{ramezani2009nash}.
Specifically, let $u_i$ denote the individual user utility.
Then, the utilitarian CUF is defined as $\sum_{i} u_i$, which favors users with higher average utility.
The Nash CUF is defined as $\sum_{i} \log(u_i)$, which balances efficiency and fairness.



\section{Policy design and analysis}

In this section, we focus on policy design and analysis for MAB-based ONL2R with both personalized and equal treatment settings.
While these two settings are different, they share some common subtasks (e.g., estimations of position preferences and arm means).
Thus, we will consider these common subtasks as preliminaries in Section~\ref{prelim} first, which paves the way for presenting our policies in Section~\ref{subsec:policies}.
Lastly, we will conduct regret analysis for our proposed policies in Section~\ref{subsec:regret}.


\subsection{Preliminaries}
\label{prelim}

{\bf 1)  Notations and Terminologies:}
Before we present our proposed bandit policy designs, we introduce some notations and terminologies as follows.
At time $t$, if the policy picks a permutation $\sigma_t$, then any arm $j \in \sigma_t$ is said to have been ``pulled'' by the agent at time $t$.
We use $T_{i,j,k}(t)$ to denote the cumulative pulling times of arm $j\in[M]$ that is offered to users of type $i\in[N]$ at position $k \in [K]$ up to time $t$, i.e., $T_{i,j,\sigma_t(j)}(t)= T_{i,j,\sigma_t(j)}(t-1) + 1$ for $i=I(t), j\in \mathrm{M}_{\sigma_t}$. 
Likewise, we use $S_{i,j,k}(t)$ to denote the cumulative reward of arm $j\in[M]$ that is clicked by users of type $i\in[N]$ at position $k\in[K]$, i.e., $S_{i,j,\sigma_t(j)}(t) = S_{i,j,\sigma_t(j)}(t-1)+ r(t)$ for $i=I(t), j=J(t)$.
We represent tensors using bold notation, e.g., $\bm{S}_{i,j}(t)=(S_{i,j,1}(t), \ldots, S_{i,j,K}(t))$.
We use the notation $\|\cdot\|_1$ to represent the $\ell^1$-norm of tensors, e.g., $\|\bm{S}_{i,j}(t) \|_1 = \sum_{k=1}^K S_{i,j,k}(t)$. 

The main statistical challenge in our model is to estimate the position preferences and arm means only by the user feedback, i.e., a sequence of the joint realizations of position preference distribution and arm reward distribution.
To this end, we present two estimators that detangle the joint realization and  estimate the unkown parameters with asymptotic confidence over time.

\begin{wrapfigure}{R}{0.45\textwidth}
\vspace{-.3in}
\begin{minipage}{0.45\textwidth}
\begin{algorithm}[H]
\small
\caption{The Position Preference Estimator $E\left( \bm{T}(t), \bm{S}(t)\right)$.}
\label{estimator}
\begin{algorithmic}[1]
	\State {\bf Input:} $\bm{T}(t), \bm{S}(t)$
    \For{all player $i$, arm $j$, and position $k$}
    	\State $\bar{v}_{i,j,k}(t) = \cfrac{S_{i,j,k}(t)/T_{i,j,k}(t)}{\sum_{l\in[K]}S_{i,j,l}(t)/T_{i,j,l}(t)}$
    \EndFor
    \For{all player $i$ and position $k$}
    	\State $\hat{\rho}_{i,k}(t) = \cfrac{1}{M} {\sum_{j\in[M]}\bar{v}_{i,j,k}(t)}$
    \EndFor
\end{algorithmic}
\end{algorithm}
\end{minipage}
\vspace{-.2in}
\end{wrapfigure}

\textbf{2) Position Preference Estimator:}
The position preference is estimated based on the fact that given a user type $i$ and an arm $j$, the value $S_{i,j,k}(t)/T_{i,j,k}(t)$ at any position $k$ is asymptotically approaching its expectation $\mu_{i,j}\cdot\rho_{i,k}$ over time.
Thus, intuitively, its normalization over all positions asymptotically removes the impact of arm mean on the position preference estimation, as stated in Algorithm~\ref{estimator}.
It is then possible to obtain a concentration on the estimated position preference, as stated in Lemma~\ref{rho} below.
Due to space limitation, the proofs of all theoretical results are relegated to the supplemental material.

\begin{restatable}{lemma}{RestateLEMMA}
\label{rho}{\em
For each user type $i\in[N]$ and position $k\in[K]$, for any constant $\epsilon\geq 0$, the position preference estimator $E(\bm{T}(t),\bm{S}(t))$ achieves a concentration bound as follows:
\begin{equation}
\label{rho_eq}
\mathbb{P}\left( |\hat{\rho}_{i,k}(t) - \rho_{i,k}|\geq\max_{j\in[M]}\sqrt{\cfrac{\epsilon\ln t}{\mu_{i,j}^2T_{i,j,k}(t)}} \right)\leq MKt^{-2\epsilon}.
\end{equation}
}
\end{restatable}

\textbf{3) Arm Mean Estimator:}
For user type $i$ and arm $j$, we note that directly estimating the arm mean $\mu_{i,j}$ by the value $\|\bm{S}_{i,j}(t)\|_1/\|\bm{T}_{i,j}(t)\|_1$ could be biased since its expectation is different from the arm mean $\mu_{i,j}$.
To address this problem, we define an asymptotically unbiased total pulling time estimator $N_{i,j}(t) = \sum_{k\in[K]} T_{i,j,k}(t)\cdot\hat{\rho}_{i,k}(t)$ for $i\in[N], j\in[M]$, which has an increment of $\hat{\rho}_{i,k}(t)$ once pulled at time $t$ and can be leveraged to estimate arm means only with the knowledge of joint realizations.
We note here that $N_{i,j}(t)$ is a random variable that depends on both the ranking history $\lbrace \bm{\sigma}(t)\rbrace_t$ and the position preference distribution.
Then, for user type $i$ and arm $j$, the asymptotically unbiased arm mean estimator can be defined as $\hat{\mu}_{i,j}(t)=\|\bm{S}_{i,j}(t-1)\|_1/N_{i,j}(t-1)$.

Following Lemma~\ref{rho}, define event $\mathcal{N}_t$ as follows: at time $t$, for user $i$ and arm $j$, there exists $\epsilon\geq 0$ such that $|N_{i,j}(t)-\bar{N}_{i,j}(t)|<\|\bm{T}_{i,j}(t)\|_1\max_{j\in[M]}\sqrt{\epsilon\ln t/\left(\mu_{i,j}^2T_{i,j,k}(t)\right)}$.
Then, we have:

\begin{restatable}{lemma}{RestateLEMMB}
\label{mu}{\em
For user type $i$ and arm $j$, denote the unbiased empirical arm means by $\hat{\mu}_{ij}(t)=\|\bm{S}_{i,j}(t-1)\|_1/N_{ij}(t-1)$.
Then, conditioned on event $\mathcal{N}_t$ and for any $\epsilon\geq 0$ we have
\begin{equation}
    \mathbb{P}\left( \big|\hat{\mu}_{i,j}(t)-\mu_{ij}\big|\geq \cfrac{\sqrt{2\epsilon\|\bm{T}_{i,j}(t)\|_1}}{N_{i,j}(t)} \hspace{4pt}\bigg|\hspace{4pt}\mathcal{N}_t\right)\leq \epsilon e^{1-\epsilon}\log t.\label{mu1}
\end{equation}
}  
\end{restatable}

\textbf{4) CUF Estimator:}
Different from personalized treatment, equal treatment policies recommend identical ranking lists to all user types, which require extra estimation of user arrival rates.
Thus, obtaining an optimal arm ranking order is typically more complicated than obtaining a decreasingly ordered arm list as in personalized treatment.
We measure the quality of a permutation by a CUF estimator.
Combining $\hat{\rho}_{i,k}(t)$ and $\hat{\mu}_{i,j}(t)$ with the fact that we can estimate the arrival rate of user type $i$ at time $t$ as $\hat{\lambda}_i(t)/\|\hat{\bm{\lambda}}(t)\|_1$, where $\hat{\lambda}_i(t)$ is the cumulative number of arrived users in type $i$ up to time $t$, we can estimate all unknown parameters $\rho_{i,k}$, $\mu_{i,j}$, and $\lambda_i$ with asymptotic confidence.
Given a ranking $\sigma$ at time $t$, we can estimate the unknown CUF $\Gamma(\sigma)$ as follows:
\begin{equation}
\label{estimate}
	\hat{\Gamma}_t(\sigma) = \hspace{-4pt}\sum_{i\in[N]} \cfrac{\hat{\lambda}_i(t)}{\|\hat{\bm{\lambda}}(t)\|_1}\cdot f\bigg( \sum_{j\in \mathrm{M}_{\sigma}} \hat{\rho}_{i,\sigma}(j)\cdot \hat{\mu}_{i,j}(t) \bigg).
\end{equation}


\subsection{MAB policy design for personalized and equal treatments}
\label{subsec:policies}
Based on the estimators presented in Section~\ref{prelim}, we are now in a position to present our policy designs for both personalized and equal treatments.

\textbf{1) GreedyRank:} GreedyRank is a greedy policy that has an increasing probability to exploit the empirical best permutation over time and a decreasing probability to explore combinations of every arm and position in a round-robin fashion.
The policy is described in Algorithm~\ref{greedy}.

\begin{algorithm}
	\caption{The GreedyRank Policy.}
    \label{greedy}
    \small
	\begin{algorithmic}[1]
    \State {\bf Input:} $\varepsilon_t$
    \State{\bf Initialization:} $\sigma_t \hspace{-2pt}=\hspace{-2pt} \big\lbrace\text{arm}\rightarrow\text{position}\hspace{-2pt}:\hspace{-2pt}[(t+k)\bmod M]+1\rightarrow k, k\hspace{-2pt}\in\hspace{-2pt}[K] \big\rbrace $ till $\min\limits_{i,j,k} S_{i,j,k}(t)\hspace{-2pt}>\hspace{-2pt}0$
    \State Mark current time $t_0=t$, and let $I_{\text{explore}}\leftarrow 1$    	\For{$t= t_0+1, \ldots$}
        \State Observe user type $I(t)$, and toss a coin with head rate of $\varepsilon_t$
        \If{head}
            \State $\sigma_t = \big\lbrace\text{arm}\rightarrow\text{position}:[(I_{explore}+k)\bmod M]+1\rightarrow k, k\in[K] \big\rbrace $
            \State $I_{\text{explore}}= (I_{\text{explore}}\bmod M) +1$
        \Else
            \State \textit{Option 1} [Personalized treatment]: $\sigma_\mu \leftarrow$ decreasingly rank arm indices by $\hat{\mu}_{I(t),j}(t)$
            \State \hspace{128pt}$\sigma_\rho \leftarrow$ decreasingly rank position indices by $\hat{\rho}_{I(t),k}(t)$
            \State \hspace{131pt}$\sigma_t=\lbrace\text{arm}\rightarrow\text{position}: \sigma_\mu(a)\rightarrow\sigma_\rho(a), a\in[K]\rbrace$
            \State \textit{Option 2} [Equal treatment]: $\sigma_t = \arg\max_{\sigma\in P_K^M}\hat{\Gamma}_t(\sigma)$
        \EndIf
        \State Observe user feedback $r(t)$, update parameters $S_{i,j,k}(t)$, $T_{i,j,k}(t)$, $\hat{\bm{\rho}}(t)$, $N_{i,j}(t)$ as stated in Section~\ref{prelim}
    \EndFor
    \end{algorithmic}
\end{algorithm}

\textbf{2) UCBRank:} Under UCBRank, the personalized treatment allows UCB-style policies to sort optimistic indices in a decreasing order and pick the corresponding first $K$ arms as permutation, while the equal treatment searches for a the permutation that maximizes the estimated CUF and a confidence interval.
To balance exploration and exploitation, we use a confidence interval derived from the McDiarmid's inequality, as described in Algorithm~\ref{ucb}.

\begin{algorithm}
    \caption{The UCBRank Policy.}
    \label{ucb}
    \small
    \begin{algorithmic}[1]
    \State {\bf Input:} $a_t$
    \State{\bf Initialization:} $\sigma_t \hspace{-2pt}=\hspace{-2pt} \big\lbrace\text{arm}\rightarrow\text{position}\hspace{-2pt}:\hspace{-2pt}[(t+k)\bmod M]+1\rightarrow k, k\hspace{-2pt}\in\hspace{-2pt}[K] \big\rbrace $ till $\min\limits_{i,j,k} S_{i,j,k}(t)\hspace{-2pt}>\hspace{-2pt}0$
    \State Mark current time $t_0=t$
    \For{$t= t_0+1, \ldots$}
        \State Observe user type $I(t)$
        \State \textit{Option 1} [Personalized treatment]: $\sigma_\mu \leftarrow$ decreasingly rank arm indices by $\hat{\mu}_{I(t),j}(t) + \cfrac{a_t\ln t}{N_{I(t),j}(t)}$
        \State \hspace{128pt}$\sigma_\rho \leftarrow$ decreasingly rank position indices by $\hat{\rho}_{I(t),k}(t)$
        \State \hspace{131pt}$\sigma_t=\lbrace\text{arm}\rightarrow\text{position}: \sigma_\mu(a)\rightarrow\sigma_\rho(a), a\in[K]\rbrace$
        \State \textit{Option 2} [Equal treatment]: $\sigma_t = \arg\max_{\sigma\in P_K^M} \left(\hat{\Gamma}_t(\sigma)+\sum_{i\in[N]}\sum_{j\in\mathrm{M}_\sigma}\cfrac{a_t\ln t}{N_{i,j}(t)} \right)$
        \State Observe user feedback $r(t)$, update parameters $S_{i,j,k}(t)$, $T_{i,j,k}(t)$, $\hat{\bm{\rho}}(t)$, $N_{i,j}(t)$ as stated in Section~\ref{prelim}
    \EndFor
    \end{algorithmic}
\end{algorithm}

One important remark regarding the time complexity of the equal treatment option in both policies is in order.
In both policies, The option of equal treatment requires solving an integer (combinatorial) optimization problem, which may be NP-hard depending on the type of the utility function $f$.
As a result, the equal treatment setting is more challenging in general than the personalized treatment setting in MAB-based ONL2R.
Also, it is easy to see that the search space of the optimization is $\mathcal{O}(M^K)$.
To this end, several interesting cases may happen:

\begin{list}{\labelitemi}{\leftmargin=1.5em \itemindent=-0.0em \itemsep=-.1em}
\vspace{-.1in}

\item[1)] {\em $K$ is Fixed and Moderate:} 
This case corresponds to a short ranking list (e.g., due to limited screen space on phones).
In this case, the search space is polynomial with respect to $M$.
Thus, the integer optimization problems in Lines~11 and 7 in Algorithms~\ref{greedy} and \ref{ucb}, respectively (referred to as IOPs) can be exactly solved by brute force search with acceptable time complexity.
However, if $M$ is fixed or $K$ is fixed but large, a brute force search is clearly too costly.

\item[2)] {\em $f$ is Linear and the $M^K$-value Is Moderate:}
This case happens if utilitarian criterion is used.
In this case, IOPs can be equivalently transformed to an integer linear program (ILP) over a probability simplex (associating each permutation with a binary variable such that these binary variables sum to 1).
Thanks to this simple structure, solving the LP relaxation of the transformed problem automatically yields a binary solution (hence the optimal ranking) in polynomial time.

\item[3)] {\em $f$ is Linear and the $M^K$-value Is Large (Exponential):}
In this case, even solving the transformed LP relaxation of the optimization problems (Lines~11 and 7 in Algorithms~\ref{greedy} and \ref{ucb}, respectively) could still be cumbersome due to the large problem size.
One solution approach is to uniformly sample with probability $p$ a subset of all possible permutations and then solve an LP relaxation only using the sampled permutations.
Clearly, as $p \rightarrow 1$, the LP solution will be arbitrarily close to the optimal solution, hence yielding a {\em polynomial-time approximation scheme} (PTAS).

\item[4)] {\em $f$ is Concave:}
This case happens if, e.g., Nash criterion ($f=\log(\cdot)$) is used.
In this case, IOPs are an integer convex optimization problem, which may be solved by state-of-the-art branch-and-bound-type (BB) optimization scheme~\cite{sierksma2015linear} to high accuracy relatively fast, if exponentially growing running time is tolerable.

\item[5)] {\em $f$ is Non-Concave:}
In this case, IOPs are an integer non-convex problem (INCP), which is the hardest.
Although one can still use branch-and-bound-type schemes in theory, the time complexity could be arbitrarily bad.
However, developing INCP algorithms is beyond the scope of this paper.
\end{list}

\subsection{Regret analysis} \label{subsec:regret}

We now present the theoretical results on the proposed policies, as well as their proof sketches.
The complete proofs are provided in the supplemental material.

\begin{restatable}{theorem}{RestateGREEDYa}(Personalized treatment with GreedyRank)
    \label{thm_greedy1}{\em
    Setting $\varepsilon_t=t^{-1/2}$, the expected regret of GreedyRank Option 1 at any time step $t$ can be bounded as follows:
    \begin{equation*}
        \mathbb{E}[R(t)]\leq 2N\sqrt{t} + \sum_{i\in[N]}\cfrac{8C_\rho MK\sqrt{\lambda_it}\ln t}{(1-1/C)\min_j\mu_{i,j}} + \mathcal{O}(1),
    \end{equation*}
    where $C, C_\rho >1$ are problem-dependent constants.
    }
\end{restatable}
\begin{proof}[Proof Sketch of Theorem~\ref{thm_greedy1}]
Based on the lemmas presented in Section~\ref{prelim}, we define two ``good'' events $\mathcal{P}_t$ and $\mathcal{U}_t$ regarding the estimated position preference and the estimated arm mean, respectively.
Then, for user type $i$ and arm $j$, conditioned on events $\mathcal{P}_t$ and $\mathcal{U}_t$, we obtain a concentration bound regarding the quantity $\hat{\mu}_{i,j}(t)\cdot\hat{\rho}_{i,\sigma_t(j)}(t)$ for any policy $\sigma_t$.
Next, we define another ``good'' event $\mathcal{F}_t$ regarding the minimum cumulative number of exploration times that GreedyRank performs up to time $t$.
Conditioned on all the defined events, we obtain the upper bound of the regret $\mathbb{E}[R(t)]$ with respect to $\varepsilon_t$, and the upper bound of $\mathbb{E}[R(t)]$ is minimized with $\varepsilon_t = t^{-1/2}$.
Finally, upper bounding the complementary of all the conditioned events finishes the proof.
\end{proof}

\begin{restatable}{theorem}{RestateUCBa}(Personalized treatment with UCBRank)
    \label{thm_ucb1}{\em
    Setting $a_t\in(2/\min\mu_{i,j}, \sqrt{t/\ln t}]$, the expected regret of UCBRank Option 1 at any time step $t$ can be bounded as follows:
    \begin{equation*}
    \mathbb{E}[R(t)] \leq \sum_{i\in[N]}\cfrac{2C_\rho MK\sqrt{\lambda_i t\ln t}}{\min_j\Delta_{i,j}} + \mathcal{O}(1).
    \end{equation*}}
\end{restatable}
\begin{proof}[Proof Sketch of Theorem~\ref{thm_ucb1}]
We define a ``bad'' event $\mathcal{E}_t$ regarding the appearance of sub-optimal permutation selection.
Then, we show that event $\mathcal{E}_t$ happens with probability zero if i) events $\mathcal{P}_t$ and $\mathcal{U}_t$ happens, ii) $a_t$ is in a proper range, and iii) each possible permutation has been sampled for enough times.
By bounding the probability of the appearance of event $\mathcal{E}_t$ for each time, and bounding the complementary of all the conditioned events as in the proof of Theorem~\ref{thm_greedy1}, we finish the proof.
\end{proof}

For the equal treatment option in both policies, to address the potential NP-Hardness challenge in solving the optimization problems (Lines~11 and 7 in Algorithms~\ref{greedy} and \ref{ucb}, respectively), we consider approximation algorithms to balance the trade-off between time complexity and optimality.
Also, compared to personalized treatment, equal treatment policies are more expensive on average due to the extra estimation of user arrival rate.

To avoid linear regret, we require the continuity assumption on the utility function $f$ as follows:
\begin{assumption}[Bi-Lipschitz Continuity] \label{assum:Lipschitz}
	{\em
	The function $f(\cdot)$ is $L_f$ bi-Lipschitz continuous, i.e., there exists a constant $L_f\geq 0$ such that for any $x_1,x_2\in[\min_{i,j,k}\lbrace\rho_{i,k} \mu_{i,j}\rbrace,\max_{i,j,k}\lbrace\rho_{i,k} \mu_{i,j}\rbrace]$, it holds that $(x_1-x_2)/L_f \leq f(x_1) - f(x_2) \leq L_f (x_1-x_2)$.
	}
\end{assumption}

\begin{restatable}{theorem}{RestateGREEDYb}(Equal Treatment with GreedyRank)
    \label{thm_greedy2}{\em
    Setting $\varepsilon_t=N t^{-1/2}$, with a $\delta_t$-approximate solution to the maximization problem in GreedyRank Option 2, the expected regret of Fair-GreedyRank at any time step $t$ can be bounded by:
    \begin{equation*}
        \mathbb{E}[R(t)]\leq 2Nt^{1\over 2} + \cfrac{8L_fC_\rho NMK}{(1-1/C)\min\mu_{i,j}}t^{1\over 2}\ln t + \sum_{s=1}^t f(1)N\delta_s + \mathcal{O}(1),
    \end{equation*}
    and for $\delta_t=\mathcal{O}(t^{-1})$, we have: $\mathbb{E}[R(t)] = \mathcal{O}\left( 8L_f NMKt^{1\over 2}\log t/ \min\mu_{i,j} \right)$.}
\end{restatable}
\begin{proof}[Proof Sketch of Theorem~\ref{thm_greedy2}]
In the equal treatment setting, we obtain a concentration on the estimated CUF $\hat{\Gamma}_t(\sigma)$ by i) the properties of the utility function $f$, and ii) the estimated user arrival rate $\hat{\lambda}_i(t)$.
Note that $\hat{\lambda}_i(t)$ can be bounded by Hoeffding's inequality.
Then, when bounding the regret upper bound in a similar manner as that in the proof of Theorem~\ref{thm_greedy1}, we additionally consider the suboptimality from the approximated solution, which causes an extra regret of at most $f(1)N\delta_t$ for each time $t$.
\end{proof}

To present the regret result of equal treatment UCBRank, we define the minimum reward gap $\Delta_{\Gamma}$ as follows:
$\Delta_{\Gamma} = \Gamma(\bm{\sigma}^*) - \max_{\bm{\sigma}\in P^M_K, \bm{\sigma}\neq\bm{\sigma}^*}\Gamma(\bm{\sigma})$.
We note that $\Delta_{\Gamma}$ only depends on the distributions of user arrival rate $\bm{\lambda}$, position preference $\bm{\rho}$, and arm mean $\bm{\mu}$.
\begin{restatable}{theorem}{RestateUCBb}(Equal Treatment with UCBRank)
    \label{thm_ucb2}{\em
    With any $\delta_t$-approximate solution to the maximization problem in UCBRank Option 2, setting $\delta=\mathcal{O}(\sqrt{\log t/t})$ and $a_t\in(2L_f/\min\mu_{i,j}, \sqrt{t/\ln t}]$, the expected regret of UCBRank at any time step $t$ can be bounded as follows:
    \begin{equation*}
    \mathbb{E}[R(t)] = \mathcal{O}\left( \cfrac{N^2MK\sqrt{t\log t}}{\Delta_{\Gamma}} \right).
    \end{equation*}}
\end{restatable}
\begin{proof}[Proof Sketch of Theorem~\ref{thm_ucb2}]
Similar as the proof of Theorem~\ref{thm_ucb1}, we define a ``bad'' event $\mathcal{E}_t$ regarding the appearance of sub-optimal permutation selection, and we show that event $\mathcal{E}_t$ happens w.p.0 if two additional conditions are satisfied: i) the estimated CUF $\hat{\Gamma}_t(\sigma_t)$ is accurate enough, which is shown in the proof of Theorem~\ref{thm_greedy2}, and ii) the suboptimality of the approximation $\delta_t$ is upper bounded.
\end{proof}



\section{Experiments}

\subsection{Experiment on synthetic data}
\begin{figure*}[t!]
    	\begin{minipage}{.48\textwidth}
    	\vspace{-6pt}
        \includegraphics[trim=50 0 35 0, width=.49\linewidth]{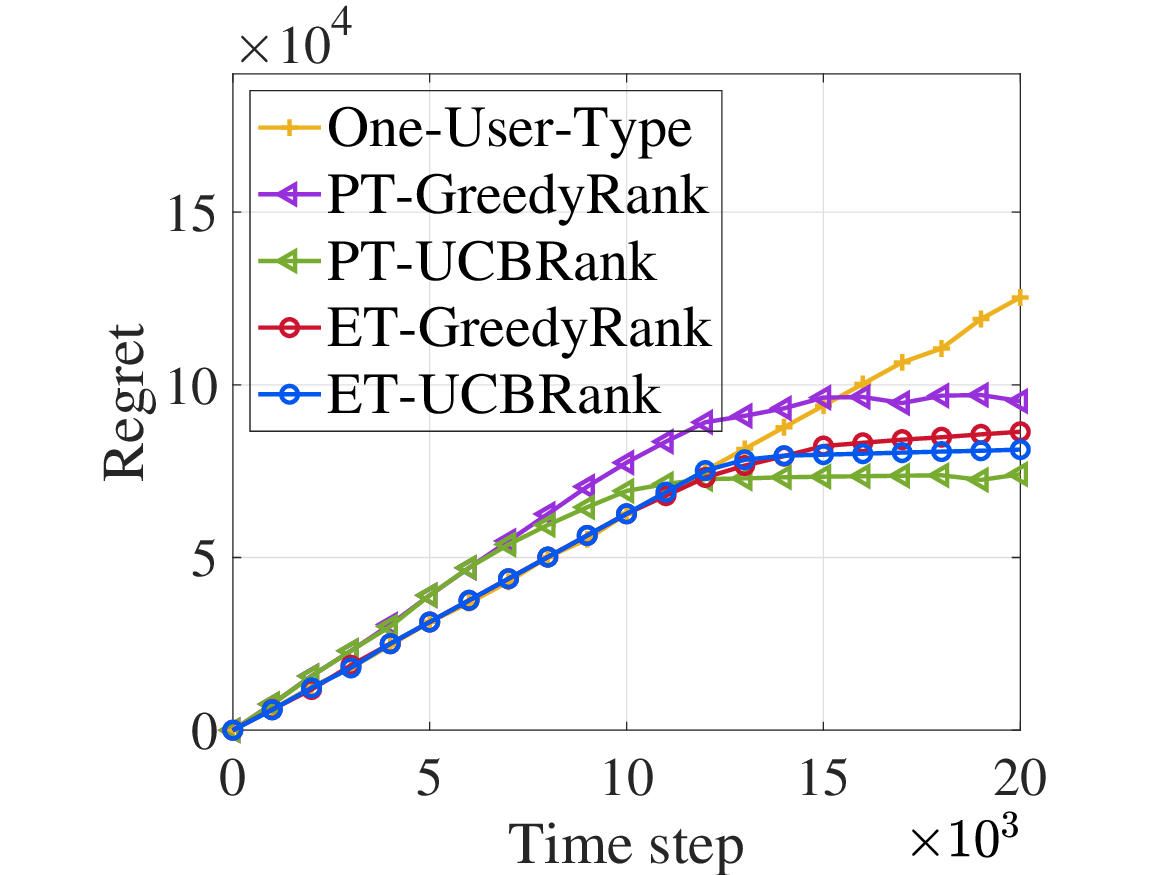}
        \hfill
        \includegraphics[trim=50 0 35 0, width=.49\linewidth]{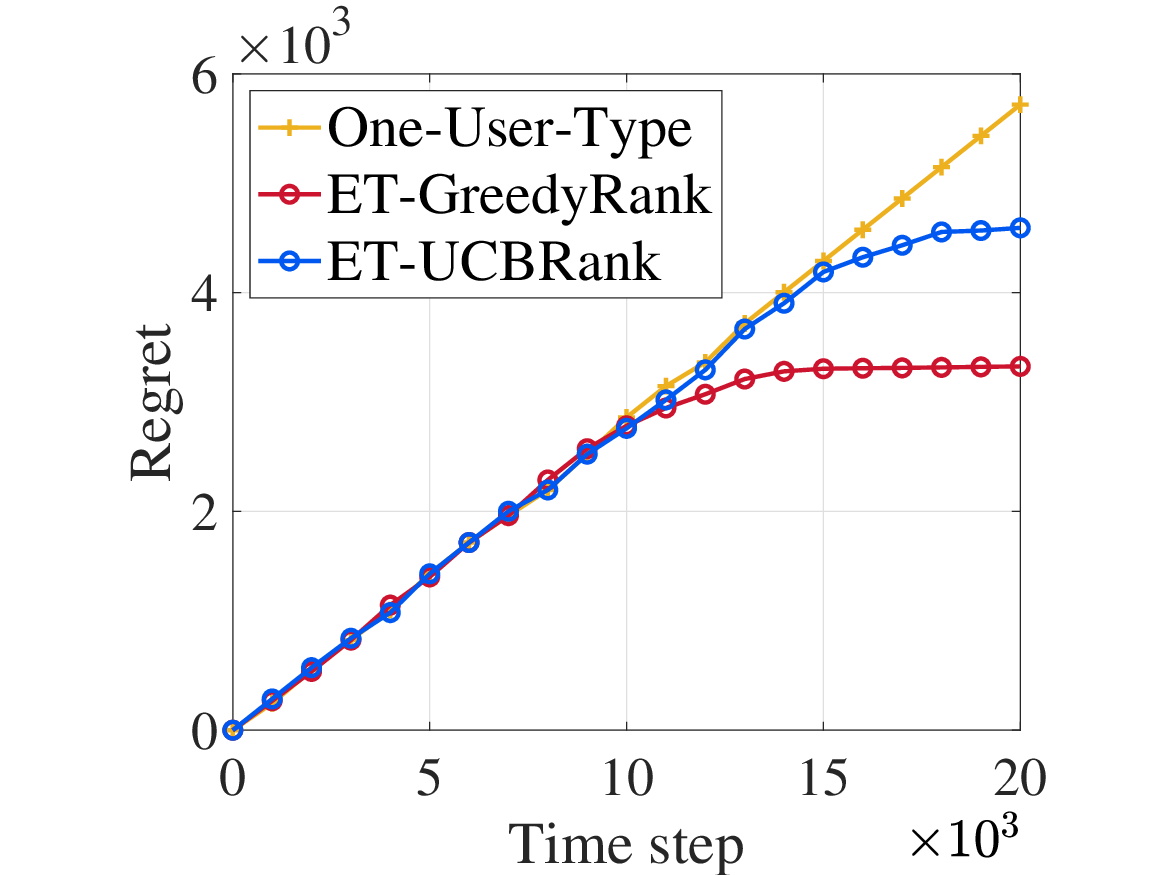}
        \caption{Baselines, personalized policies (left), equal treatment policies in utilitarian CUF (left) and Nash CUF (right) on synthetic dataset.}
        \label{beta}
        \end{minipage}
        \hfill
        \begin{minipage}{.48\textwidth}
        \vspace{-6pt}
        \includegraphics[trim=50 0 35 0, width=.49\linewidth]{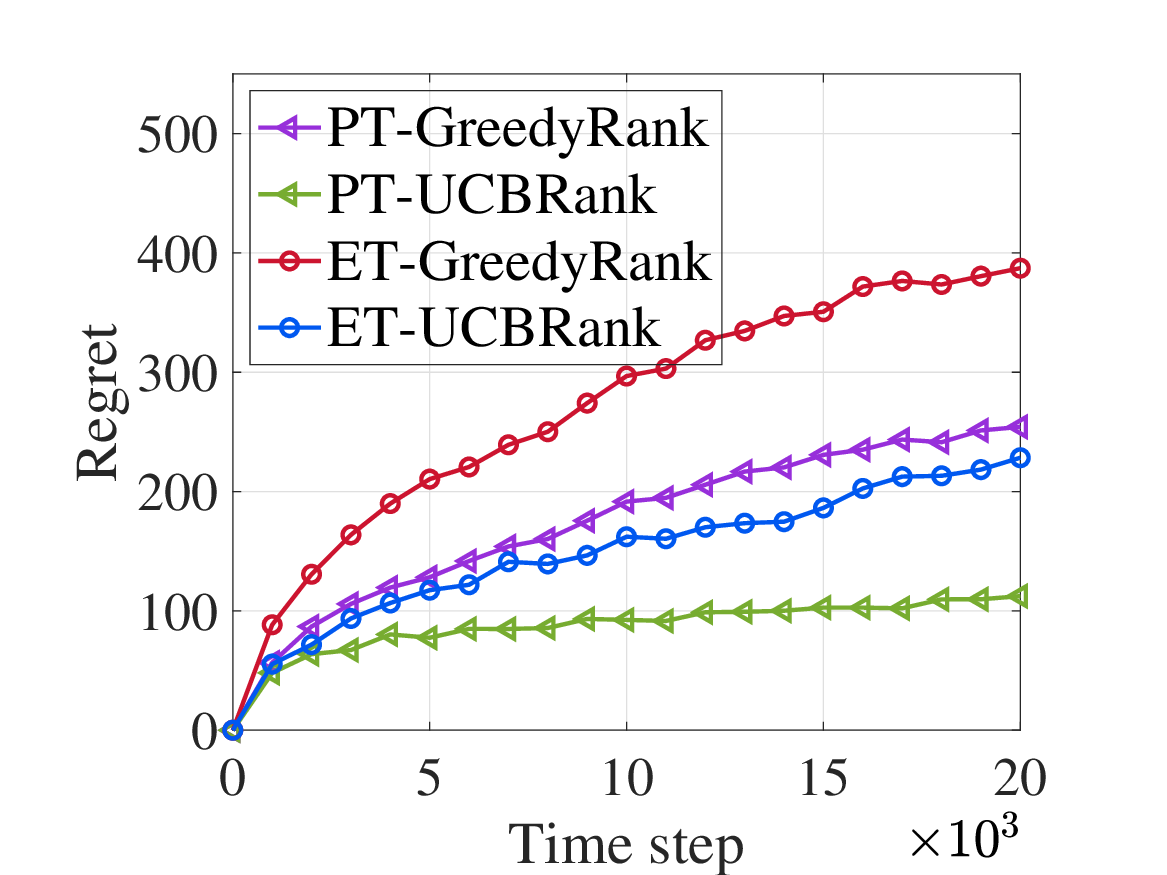}
        \hfill
        	\includegraphics[trim=40 8 30 0, width=.49\linewidth]{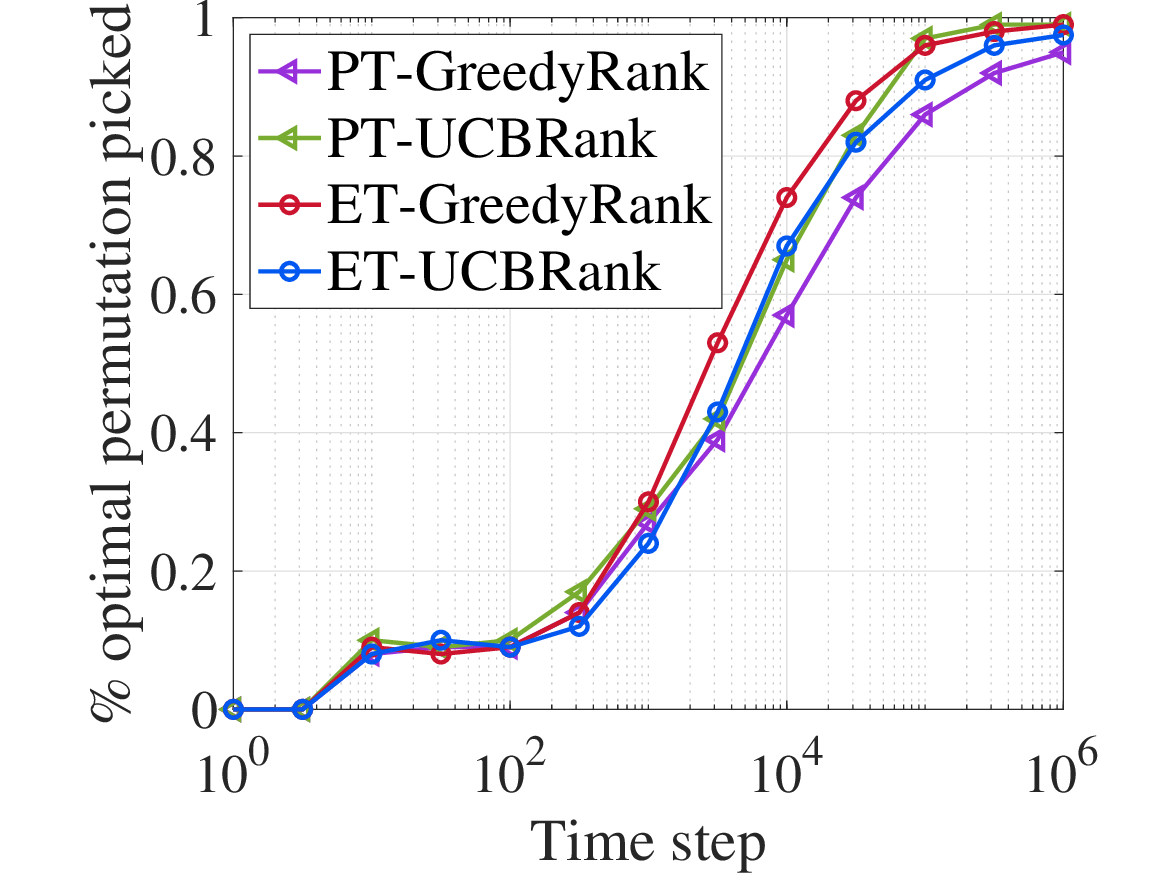}
        	\caption{Average regret of proposed policies (left), and the optimal action rate of proposed policies (right) on real-world dataset.}
        	\label{real}
        	\end{minipage}
    \label{fig:experiment}
    \vskip -0.21in
    \end{figure*}

We first conduct experiment on a synthetic dataset, where $N=3$, $M=20$, and $K=4$.
We relax the assumption of Bernoulli distributed arm means, and replace it by the Beta distribution, which allows discrete-valued reward with a scaled-up arm expectation (this will increase the regret while it helps the estimation with a large problem size in a finite-time horizon).
All the system parameters are randomly sampled.
We set $\varepsilon_t=a_t=1$ for personalized treatment GreedyRank (PT-GreedyRank) and UCBRank (PT-UCBRank), respectively.
We set $\varepsilon_t=a_t=5$ for equal treatment GreedyRank (ET-GreedyRank) and UCBRank (ET-UCBRank), respectively.
Since there is no existing algorithm that works in our context, we set a baseline with a idea similar to most existing algorithms: the baseline runs the UCB algorithm that shares the same confidence interval as ours in PT-UCBRank, while the baseline does not distinguish different user types and treats all users as one type.
We present the results in Fig.~\ref{beta}.
The curves confirm our analysis that all proposed policies are sub-linear in regret, and the performance of each policy also depends on the system parameters and policy parameters, e.g., ET-GreedyRank is optimal in utilitarian CUF but not in Nash CUF.

\subsection{Experiment on real-world data}

We use the dataset provided for KDD Cup 2012 track 2 \cite{kddtrack2}, which is about advertisements shown alongside search results in a search engine owned by Tencent.
The users in the dataset are numbered in millions, and are provided with demographics information, e.g., gender.
Ads are displayed in a position (1, 2, or 3) with a binary reward (click or not).
Since ads are rarely displayed in position 3, which results in a lack of data, so we focus on two positions (1 and 2).
We pick the top 5 ads with high frequency and present the statistical information in Table~\ref{t1}.
We set $\varepsilon_t=a_t=.25$ for PT-GreedyRank and PT-UCBRank, respectively.
We set $\varepsilon_t=a_t=.5$ for ET-GreedyRank and ET-UCBRank, respectively.
We present the results in Fig.~\ref{real}.
The results show that all the proposed policies find the optimal permutations over time.
Although the rewards are synthetic, this experiment is still realistic since the values of all other parameters are extracted from the real world.

\begin{minipage}[c]{0.48\textwidth}
\centering
\vskip 0.22in
\resizebox{.99\columnwidth}{!}{
\begin{tabular}{lccccccr}
\toprule
    &\multicolumn{5}{c}{Arm Mean} & \multicolumn{2}{c}{Position Bias} \\
\cmidrule(lr){2-6}\cmidrule(lr){7-8}
Gender & $1$ & $2$ & $3$ & $4$ & $5$ & $1$ & $2$ \\
\midrule
Male   & $.357$ & $.471$ & $.604$ & $.808$ & $.564$ & $.323$ & $.677$ \\
Female & $.247$ & $.327$ & $.491$ & $.49$ & $.303$ & $.416$ & $.584$ \\
\midrule
\multicolumn{8}{l}{Arrival Rate: $.52$(M) : $.48$(F)} \\
\bottomrule
\end{tabular}
}
\captionof{table}{Statistics on the dataset, including two user types (male and female), user arrival rate, arm means and position preference (bias).}
\label{t1}
\end{minipage}
\hfill
\begin{minipage}[c]{0.48\textwidth}
\centering
\resizebox{.99\columnwidth}{!}{
\begin{tabular}{lccccccr}
\toprule
   &   & \multicolumn{2}{c}{$t=3\cdot 10^5$} & \multicolumn{2}{c}{$t=6\cdot 10^5$} \\
   \cmidrule(lr){3-4}\cmidrule(lr){5-6}
   &   & regret & \makecell{running \\ time$/s$} & regret & \makecell{running \\ time$/s$} \\
\midrule
\multirow{2}{*}{GreedyRank\hspace{-10pt}} & approx. & $512$ & $33$ & $581$ & $68$ \\
   & opt.      & $387$ & $41$ & $461$ & $87$ \\
\midrule
\multirow{2}{*}{UCBRank}    & approx. & $314$ & $75$ & $326$ & $162$ \\
   & opt.      & $238$ & $82$ & $249$ & $188$ \\
\bottomrule
\end{tabular}
}
\captionof{table}{Comparison of equal treatment policies with approximated solution and optimal solution under utilitarian CUF.}
\label{t2}
\vskip -0.21in
\end{minipage}

We also compare the case where approximation algorithms are used in equal treatment ranking.
we use a PTAS for utilitarian CUF maximization in both policies: given a ratio $\delta_t$ that gets close to one over time, randomly sample $\delta_tn(P^M_K)$ permutations and find the optimal in the samples.
If the utility function $f$ is linear, it can be easily shown that this strategy is PTAS.
The experiment on real-world dataset runs on CPU configured by Apple M1 with 8-core and 3.2 GHz, with 16 GB main memory.
The results in Table~\ref{t2} confirm our analysis on the tradeoff between regret and time complexity.

\section{Conclusion}
This paper studied personalized and equal treatment rankings in position-based ONL2R recommendations.
All proposed policies achieve sub-linear regrets without the information of user arrival rate, position preference, and arm means.
Potential future research directions include extending the position-based click model to other practical click models, e.g., cascade models, or further relaxing the model assumptions in this paper, such as theoretical analysis of Beta distribution of arm means, under which our policies still reach sub-linear regret practically as shown in the experiment results.


\bibliography{paper}
\bibliographystyle{plain}

\newpage
{\bf\LARGE Supplementary Material}

\section{Proof of Lemma~\ref{rho}}
\RestateLEMMA*
\begin{proof}
In Algorithm~\ref{estimator}, we denote a random vector $\bm{v}_{i,j}(t) = \bm{S}_{i,j}(t)/\bm{T}_{i,j}(t)$ for user $i$ and arm $j$.
Given that each entry of vectors $\bm{S}_{i,j}(t)$ and $\bm{T}_{i,j}(t)$ is unbiased, and the arriving user at time $t$ views exactly one position, for any time step $t$ with arriving user $I(t)=i$ and arm $j\in m(t)$, we have $\mathbb{E}[v_{i,j,k}(t)-v_{i,j,k}(t-1)]=\mu_{i,j}\rho_{i,k}$ for position $k$.
Then, by Hoeffding's Inequality, for any $\epsilon\geq 0$ we have
\begin{equation}
\label{rho2}
    \mathbb{P}\left(| v_{i,j,k}(t) - \mu_{i,j}\rho_{i,k}| \geq\sqrt{\cfrac{\epsilon\ln t}{T_{i,j,k}(t)}} \right)\leq t^{-2\epsilon}, i\in[N], j\in[M], k\in[K].
\end{equation}
Summing $v_{i,j,k}(t)$ in (\ref{rho2}) over $k$, by union bound, we have
\begin{equation*}
    \mathbb{P}\left(\sum_{k\in[K]}\left(| v_{i,j,k}(t) - \mu_{i,j}\rho_{i,k}|\right) \geq K\sqrt{\cfrac{\epsilon\ln t}{T_{i,j,k}(t)}} \right) \hspace{-2pt}=\hspace{-2pt} \mathbb{P}\left(| \bm{v}_{i,j}(t) - \mu_{i,j}| \geq K\sqrt{\cfrac{\epsilon\ln t}{T_{i,j,k}(t)}} \right) \hspace{-3pt}\leq\hspace{-3pt} Kt^{-2\epsilon}.
\end{equation*}
To prove Eq.~(\ref{rho_eq}), we start from one direction of the inequality. By setting $\epsilon=0$, it is obvious that $\mathbb{P}(\bm{v}_{i,j}(t)\leq\mu_{i,j})\leq 1$.
Then, for any position $k\in[K]$, we have
\begin{align*}
    &\hspace{13pt}\mathbb{P}\left( \cfrac{1}{\bm{v}_{i,j}(t)} \geq \cfrac{1}{\mu_{i,j}}, \hspace{3pt} v_{i,j,k}(t)\geq \mu_{i,j}\rho_{i,k}+\sqrt{\cfrac{\epsilon\ln t}{T_{i,j,k}(t)}} \right)\\
    &= \mathbb{P}\left( \cfrac{v_{i,j,k}(t)}{\bm{v}_{i,j}(t)} \geq \rho_{i,k}+\sqrt{\cfrac{\epsilon\ln t}{\mu_{i,j}^2T_{i,j,k}(t)}} \right)\\
    &\leq Kt^{-2\epsilon}.
\end{align*}
Averaging the term $\cfrac{v_{i,j,k}(t)}{\bm{v}_{i,j}(t)}$ over arms $j\in[M]$, by union bound we have
\begin{align*}
    &\hspace{15pt}\mathbb{P}\left( \cfrac{1}{M}\sum_{j\in[M]}\cfrac{v_{i,j,k}(t)}{\bm{v}_{i,j}(t)} \geq \rho_{i,k}+\cfrac{1}{M}\sum_{j\in[M]}\sqrt{\cfrac{\epsilon\ln t}{\mu_{i,j}^2T_{i,j,k}(t)}} \right)\leq MKt^{-2\epsilon},\\
    &\Rightarrow \mathbb{P}\left( \hat{\rho}_{i,k}\geq \rho_{i,k}+\cfrac{1}{M}\sum_{j\in[M]}\sqrt{\cfrac{\epsilon\ln t}{\mu_{i,j}^2T_{i,j,k}(t)}} \right)\leq MKt^{-2\epsilon},\\
    &\Rightarrow \mathbb{P}\left( \hat{\rho}_{i,k}\geq \rho_{i,k}+\max_{j\in[M]}\sqrt{\cfrac{\epsilon\ln t}{\mu_{i,j}^2T_{i,j,k}(t)}} \right)\leq MKt^{-2\epsilon}.
\end{align*}
Thus, we obtain one direction of Eq.~(\ref{rho_eq}).
Similarly, to prove the reversed direction, for any position $k\in[K]$, we have
\begin{align*}
    &\hspace{13pt}\mathbb{P}\left( \cfrac{1}{\bm{v}_{i,j}(t)} \leq \cfrac{1}{\mu_{i,j}}, \hspace{3pt} v_{i,j,k}(t)\leq \mu_{i,j}\rho_{i,k}-\sqrt{\cfrac{\epsilon\ln t}{T_{i,j,k}(t)}} \right)\\
    &= \mathbb{P}\left( \cfrac{v_{i,j,k}(t)}{\bm{v}_{i,j}(t)} \leq \rho_{i,k}-\sqrt{\cfrac{\epsilon\ln t}{\mu_{i,j}^2T_{i,j,k}(t)}} \right)\\
    &\leq Kt^{-2\epsilon}.
\end{align*}
Averaging the term $\cfrac{v_{i,j,k}(t)}{\bm{v}_{i,j}(t)}$ over arms $j\in[M]$, we obtain the following
\begin{align*}
    &\hspace{15pt}\mathbb{P}\left( \cfrac{1}{M}\sum_{j\in[M]}\cfrac{v_{i,j,k}(t)}{\bm{v}_{i,j}(t)} \leq \rho_{i,k}-\cfrac{1}{M}\sum_{j\in[M]}\sqrt{\cfrac{\epsilon\ln t}{\mu_{i,j}^2T_{i,j,k}(t)}} \right)\leq MKt^{-2\epsilon},\\
    &\Rightarrow \mathbb{P}\left( \hat{\rho}_{i,k}\leq \rho_{i,k}-\cfrac{1}{M}\sum_{j\in[M]}\sqrt{\cfrac{\epsilon\ln t}{\mu_{i,j}^2T_{i,j,k}(t)}} \right)\leq MKt^{-2\epsilon},\\
    &\Rightarrow \mathbb{P}\left( \hat{\rho}_{i,k}\leq \rho_{i,k}-\max_{j\in[M]}\sqrt{\cfrac{\epsilon\ln t}{\mu_{i,j}^2T_{i,j,k}(t)}} \right)\leq MKt^{-2\epsilon}.
\end{align*}
Thus, we obtain both directions of Eq.~(\ref{rho_eq}).
\end{proof}

\section{Proof of Lemma~\ref{mu}}
\RestateLEMMB*

We will leverage the following proposition in the proof.
\begin{proposition}[\cite{lagree2016multiple}, Proposition~8]
\label{prop1}
Given user $i\in[N]$ and arm $j\in[M]$, for any $\epsilon\geq 0$, we have
\begin{equation}
    \mathbb{P}\left( \big\arrowvert\bar{\mu}_{ij}(t)-\mu_{ij}\big\arrowvert\geq \cfrac{\sqrt{{\epsilon\over 2}\|\bm{T}_{i,j}(t)\|_1}}{\bar{N}_{ij}(t)} \right)\leq \epsilon e^{1-\epsilon}\log t.
\end{equation}
\end{proposition}

\begin{proof}
Assume that the position probabilities $\bm{\rho}$ are known, then we can replace Line~16 in Algorithm~\ref{greedy} by $\bar{N}_{i,j}(t)\leftarrow \bar{N}_{i,j}(t-1) + \rho_{i,\sigma_t(j)}$, and denote unbiased empirical means by $\bar{\mu}_{i,j}(t)=\sum_kS_{i,j,k}(t-1)/\bar{N}_{i,j}(t-1)$.
Formally, define $\bar{N}_{i,j}(t)=\sum_{s=1}^t \rho_{i,\sigma_{s-1}(j)}\cdot\mathrm{1}\lbrace j\in \mathrm{M}_{\sigma_s}\rbrace$, and define $N_{i,j}(t)=\sum_{s=1}^t \hat{\rho}_{i,\sigma_{s-1}(j)}\cdot\mathrm{1}\lbrace j\in \mathrm{M}_{\sigma_s}\rbrace$, where $\mathrm{1}\lbrace X\rbrace$ is an indicator of event $X$.
Then we have:
\begin{equation*}
    |N_{i,j}(t) - \bar{N}_{i,j}(t)|/\|\bm{T}_{i,j}(t)\|_1 = \cfrac{1}{\|\bm{T}_{i,j}(t)\|_1}\sum_{s=1}^t \left[\left|\hat{\rho}_{i,\sigma_{s-1}(j)} - \rho_{i,\sigma_{s-1}(j)}\right|\cdot\mathrm{1}\lbrace j\in \mathrm{M}_{\sigma_s}\rbrace\right].
\end{equation*}
    Define a ``good'' event $\mathcal{N}_t$ as follows: at time $t$, for any user $i\in[N]$ and any arm $j\in[M]$, there exists $\epsilon\geq 0$ such that $|N_{i,j}(t)-\bar{N}_{i,j}(t)|<\|\bm{T}_{i,j}(t)\|_1\max_{j\in[M]}\sqrt{\epsilon\ln t/\left(\mu_{i,j}^2T_{i,j,k}(t)\right)}$.
Combining with Lemma~\ref{rho}, we obtain that $\mathbb{P}\left(\mathcal{N}_t^C\right)$ is upper bounded by
\begin{equation*}
    \cfrac{1}{\|\bm{T}_{i,j}(t)\|_1}\sum_{s=1}^t \left[\mathrm{1}\lbrace j\in \mathrm{M}_{\sigma_s}\rbrace\cdot\mathbb{P}\left( \left|\hat{\rho}_{i,\sigma_{s-1}(j)} - \rho_{i,\sigma_{s-1}(j)}\right|\geq \max_{j\in[M]}\sqrt{\cfrac{\epsilon\ln t}{\mu_{i,j}^2T_{i,j,k}(t)}}\right)\right]\leq MKt^{-2\epsilon}.
\end{equation*}
Then, with Proposition~\ref{prop1}, we have
\begin{align*}
    &\hspace{13pt}\mathbb{P}\left( \bar{\mu}_{ij}(t)-\mu_{ij}\geq \cfrac{\sqrt{{\epsilon\over 2}\|\bm{T}_{i,j}(t)\|_1}}{\bar{N}_{ij}(t)} \right)\\
    &= \mathbb{P}\left( \hat{\mu}_{ij}(t)-\cfrac{\mu_{ij}\bar{N}_{ij}(t)}{N_{i,j}(t)}\geq \cfrac{\sqrt{{\epsilon\over 2}\|\bm{T}_{i,j}(t)\|_1}}{N_{ij}(t)} \right)\\
    &= \mathbb{P}\left( \hat{\mu}_{ij}(t)-\mu_{ij}\geq \cfrac{\mu_{i,j}(\bar{N}_{i,j}(t)-N_{i,j}(t)) + \sqrt{{\epsilon\over 2}\|\bm{T}_{i,j}(t)\|_1}}{N_{ij}(t)} \right)\\
    &\leq \epsilon e^{1-\epsilon}\log t.
\end{align*}
Replacing $\left(\bar{N}_{i,j}(t)-N_{i,j}(t)\right)$ by $\|\bm{T}_{i,j}(t)\|_1\max_{j\in[M]}\sqrt{\cfrac{\epsilon_1\ln t}{\mu_{i,j}^2T_{i,j,k}(t)}}$ with $\epsilon_1=\epsilon/\ln t$, conditioned on $\mathcal{N}_t$, we have
\begin{align*}
&\hspace{15pt}\mathbb{P}\left( \hat{\mu}_{ij}(t)-\mu_{ij}\geq \cfrac{\sqrt{2\epsilon\|\bm{T}_{i,j}(t)\|_1}}{N_{ij}(t)}\hspace{4pt}\bigg\vert\hspace{4pt}\mathcal{N}_t \right)\\
&\leq \mathbb{P}\left( \hat{\mu}_{i,j}(t)-\mu_{i,j}\geq \cfrac{\mu_{i,j}(\bar{N}_{i,j}(t)-N_{i,j}(t))+\sqrt{{\epsilon\over 2}\|\bm{T}_{i,j}(t)\|_1}}{N_{i,j}(t)} \right)\\
&\leq \epsilon e^{1-\epsilon}\log t.
\end{align*}
Thus, we obtain one direction of Eq.~(\ref{mu1}).
Similarly, for the reversed direction, we have
\begin{align*}
&\hspace{13pt}\mathbb{P}\left( \mu_{ij}(t)-\bar{\mu}_{ij}\geq \cfrac{\sqrt{{\epsilon\over 2}\|\bm{T}_{i,j}(t)\|_1}}{\bar{N}_{ij}(t)} \right)\\
&= \mathbb{P}\left( \cfrac{\mu_{ij}(t)\bar{N}_{i,j}(t)}{N_{i,j}(t)}-\hat{\mu}_{ij}\geq \cfrac{\sqrt{{\epsilon\over 2}\|\bm{T}_{i,j}(t)\|_1}}{N_{ij}(t)} \right)\\
&= \mathbb{P}\left( \mu_{ij}(t)-\hat{\mu}_{ij}\geq \cfrac{\mu_{i,j}(N_{i,j}(t)-\bar{N}_{i,j}(t)) + \sqrt{{\epsilon\over 2}\|\bm{T}_{i,j}(t)\|_1}}{N_{ij}(t)} \right)\\
&\leq \epsilon e^{1-\epsilon}\log t.
\end{align*}
Replacing $\left(N_{i,j}(t)-\bar{N}_{i,j}(t)\right)$ by $\|\bm{T}_{i,j}(t)\|_1\max_{j\in[M]}\sqrt{\cfrac{\epsilon_1\ln t}{\mu_{i,j}^2T_{i,j,k}(t)}}$ with $\epsilon_1=\epsilon/\ln t$, conditioned on $\mathcal{N}_t$, we have
\begin{equation}	
\label{mu_concentration}
\mathbb{P}\left( \mu_{ij}(t)-\hat{\mu}_{ij}\geq \cfrac{\sqrt{2\epsilon\|\bm{T}_{i,j}(t)\|_1}}{N_{i,j}(t)}\hspace{4pt}\bigg\vert\hspace{4pt}\mathcal{N}_t \right)
\leq \epsilon e^{1-\epsilon}\log t.
\end{equation}

\end{proof}

\section{Proof of Theorem~\ref{thm_greedy1}}
\RestateGREEDYa*
We will leverage the following lemma and proposition in the proof.

\begin{proposition}[McDiarmid's Inequality]
\label{McDiarmid}
Let $X_1, \ldots, X_n$ be independent (not necessarily identical in distribution) random variables. Let $f: \mathcal{X}_1\times\cdots\mathcal{X}_n\rightarrow\mathbb{R}$ be any function with the $(c_1, \ldots, c_n)$-bounded difference property: for every $i=1, \ldots, n$ and every $(x_1, \ldots, x_n), (x'_1, \ldots, x'_n)\in \mathcal{X}_1\times\cdots\mathcal{X}_n$ that differ only in the $i$-th coordinate $(x_j=x'_j$ for all $j\neq i)$, we have $|f(x_1, \ldots, x_n) - f(x'_1, \ldots, x'_n)|\leq c_i$. Then , for any $t>0$, we have
\begin{equation*}
    \mathbb{P}\left( \left|f(x_1, \ldots, x_n) - \mathbb{E}[f(x_1, \ldots, x_n)]\right| \geq t \right) \leq \exp\left( -\cfrac{2t^2}{\sum_{i=1}^n c_i^2} \right).
\end{equation*}
\end{proposition}

\begin{lemma}
\label{lambda}
At time $t$, for any user $i\in[N]$ and any $\epsilon\geq 0$, the estimated user arrival rate $\hat{\lambda}_i(t)$ in Greedy Ranking satisfies the following
\begin{equation*}
\mathbb{P}\left( t\cdot\left| \hat{\lambda}_i(t) - \lambda_i \right|\geq \epsilon \right) \leq \exp\left( -2t\epsilon^2 \right).
\end{equation*}
\end{lemma}

\begin{proof}
We start from evaluating the initialization phase.
Since the initialization performs a round-robin sampling, then, at time $t>1$, for each user $i$, arm $j$, position $k$, we have
\begin{equation*}
    \mathbb{E}[S_{i,j,k}(t) - S_{i,j,k}(t-1)] = \cfrac{\lambda_i\mu_{i,j}}{M}
\end{equation*}
By the definition of time step $t_0$, we have
\begin{equation*}
    \mathbb{E}[t_0]\leq \sum_{i\in[N]}\sum_{j\in[M]}\cfrac{MK}{\lambda_i\mu_{i,j}}.
\end{equation*}
Next, we analyze the exploration.
Denote the cumulative number of exploration times that policy Greedy Ranking performs up to time $t$ by $\xi_t$, then we observe the following
\begin{equation*}
\mathbb{E}[\xi_t]=\sum_{s\leq t}\varepsilon_s\geq t\cdot\varepsilon_t.
\end{equation*}
Assume that there exists a constant $C>1$ such that $\varepsilon_t\geq Ct^{-1/2}$.
Then, by Hoeffding's Inequality, we obtain
\begin{equation*}
    \mathbb{P}\left(\cfrac{1}{t}\cdot\bigg| \mathbb{E}[\xi_t] - \xi_t \bigg| \geq \cfrac{\varepsilon_t}{C}\right)\leq \exp\left(-\cfrac{2t\varepsilon_t^2}{C^2}\right)
\end{equation*}
By the relations $\mathbb{E}[\xi_t]\geq t\varepsilon_t$ and $\varepsilon_t\geq Ct^{-1/2}$, we have
\begin{equation*}
    \mathbb{P}\left( t\varepsilon_t - \xi_t \geq \cfrac{t\varepsilon_t}{C}\right)\leq \exp\left(-\cfrac{2t\varepsilon_t^2}{C^2}\right)\leq \exp\left( -2t^{1/2} \right).
\end{equation*}
Define an event $\mathcal{F}_t$ regarding $\xi_t$ as follows: at time $t$, $\xi_t\geq (1-1/C)\cdot t\varepsilon_t$, and we have
\begin{equation*}
    \mathbb{P}(\mathcal{F}_t^C) = \mathbb{P}\left(\xi_t\leq t\varepsilon_t - \cfrac{t\varepsilon_t}{C}\right)\leq\exp\left( -2t^{1/2} \right) < \exp(-2\ln t) = {1\over t^2}.
\end{equation*}
In policy GreedyRank, we perform round robin over users and arms during exploration, thus, conditioned on event $\mathcal{F}_t$, for each user $i\in[N]$ and each arm $j\in[M]$ we have 
\begin{equation*}
    \|\bm{T}_{i,j}(t)\|_1\geq {\xi_t\over NM} \geq {(1-1/C)\cdot t\varepsilon_t\over NM}.
\end{equation*}

The analysis of exploitation requires a high probability bound of $|\hat{\mu}_{i,j}(t)\cdot\hat{\rho}_{i,\sigma_t(j)}(t) - \mu_{i,j}\cdot\rho_{i,\sigma_t(j)}|$.
We note that given a policy $\lbrace \sigma_t\rbrace_t$, the expected value of $\hat{\mu}_{i,j}(t)\cdot\hat{\rho}_{i,\sigma_t(j)}(t)$ depends on the policy, and may change over time.
In other words, the expected value of $\hat{\mu}_{i,j}(t)\cdot\hat{\rho}_{i,\sigma_t(j)}(t)$ is some function of a policy-related sequence of samples.
Thus, to bound the deviation of $\hat{\mu}_{i,j}(t)\cdot\hat{\rho}_{i,\sigma_t(j)}(t)$ to its expectation, we fix a user type $i$ and an arm $j$, and we use McDiarmid's inequality as stated in Proposition~\ref{McDiarmid}.
Define an event $\mathcal{P}_t$ regarding $\hat{\bm{\rho}}(t)$ as follows: at time $t$, for any user $i\in[N]$ and any position $k\in[K]$, it holds that $|\hat{\rho}_{i,k}(t)-\rho_{i,k}(t)|<\max_{j\in[M]}\sqrt{2\ln t/(\mu_{i,j}^2T_{i,j,k}(t))}$.
Define an event $\mathcal{U}_t$ regarding $\hat{\bm{\mu}}(t)$ as follows: at time $t$, for any user $i$ and any arm $j$, it holds that $|\hat{\mu}_{i,j}(t) - \mu_{i,j}|<\sqrt{2}/N_{i,j}(t)$.
Then, given any policy $\lbrace \sigma_t\rbrace_t$, at time $t$, conditioned on event $\mathcal{P}_t\cap \mathcal{U}_t$, for any user $i$ and any arm $j$, the random variable $\hat{\mu}_{i,j}(t)\cdot\hat{\rho}_{i,\sigma_t(j)}(t)$ can change by at most $\sqrt{4\ln t}/\left(N_{i,j}(t)\min_{j}\sqrt{\mu_{i,j}^2T_{i,j,k}(t)}\right)$.
By Proposition~\ref{McDiarmid}, we have
\begin{equation*}
    \mathbb{P}\left( \left|\hat{\mu}_{i,j}(t)\cdot\hat{\rho}_{i,\sigma_t(j)}(t) - \mu_{i,j}\cdot\rho_{i,\sigma_t(j)}\right| \geq \cfrac{2\ln t}{\mu_{i,j}N_{i,j}(t)} \hspace{4pt}\bigg|\hspace{4pt} \mathcal{P}_t, \mathcal{U}_t \right) \leq t^{-2},
\end{equation*}
where $\mathbb{P}(\mathcal{P}_t^C)\leq MKt^{-4}$, and $\mathbb{P}(\mathcal{U}_t^C)\lesssim t^{-1/2}\exp(1-t^{-1/2})\log t$.
By union bound, we obtain
\begin{equation*}
    \mathbb{P}\left( \left|\sum_{j\in\mathrm{M}_{\sigma}}\hat{\mu}_{i,j}(t)\cdot\hat{\rho}_{i,\sigma_t(j)}(t) - \sum_{j\in\mathrm{M}_{\sigma}}\mu_{i,j}\cdot\rho_{i,\sigma_t(j)}\right| \geq \sum_{j\in\mathrm{M}_{\sigma}}\cfrac{2\ln t}{\mu_{i,j}N_{i,j}(t)} \hspace{4pt}\Bigg|\hspace{4pt} \mathcal{P}_t, \mathcal{U}_t \right) \leq Kt^{-2}.
\end{equation*}
Now, denote $\Gamma_i(\sigma) = \sum_{j\in\mathrm{M}_\sigma}\mu_{i,j}\cdot \rho_{i,\sigma(j)}$, and denote $\hat{\Gamma}_i^t(\sigma)=\sum_{j\in\mathrm{M}_\sigma}\hat{\mu}_{i,j}(t)\cdot \hat{\rho}_{i,\sigma(j)}(t)$, by union bound we have
\begin{align*}
    & \hspace{13pt}\mathbb{P}\left( \left|\sum_{j\in\mathrm{M}_{\sigma}}\hat{\mu}_{i,j}(t)\cdot\hat{\rho}_{i,\sigma_t(j)}(t) - \sum_{j\in\mathrm{M}_{\sigma}}\mu_{i,j}\cdot\rho_{i,\sigma_t(j)}\right|\geq \sum_{j\in\mathrm{M}_{\sigma}}\cfrac{2\ln t}{\mu_{i,j}N_{i,j}(t)}\hspace{4pt}\Bigg|\hspace{4pt}\mathcal{P}_t, \mathcal{U}_t \right)\\
    & =\mathbb{P}\left( \left| \hat{\Gamma}_i^t(\sigma) - \Gamma_i(\sigma) \right|\geq\sum_{j\in\mathrm{M}_\sigma}\cfrac{2\ln t}{\mu_{i,j}N_{i,j}(t)} \hspace{4pt}\Bigg|\hspace{4pt} \mathcal{P}_t, \mathcal{U}_t \right)\\
    & \leq Kt^{-2}.
\end{align*}

Define an event $\mathcal{Y}_t^i$ regarding the estimated CUF $\hat{\Gamma}_i^t(\sigma_t)$ as follows: for any policy $\lbrace \sigma_t\rbrace_t$ and any user type $i$, it holds that $|\hat{\Gamma}_i(\sigma_t) - \Gamma_i(\sigma_t)| < \sum_{j\in\mathrm{M}_{\sigma_t}}2\ln t/(\mu_{i,j}N_{i,j}(t))$.
Conditioned on event $\mathcal{F}_t\cap \mathcal{Y}_t^1\cap \ldots\cap\mathcal{Y}_t^N$, we have
\begin{align*}
\Gamma_i(\sigma^*) - \Gamma_i(\sigma_t)
&= \left[ \Gamma_i(\sigma^*) - \hat{\Gamma}_i^t(\sigma^*) \right] + \left[ \hat{\Gamma}_i^t(\sigma^*) - \hat{\Gamma}_i^t(\sigma_t)\right] + \left[ \hat{\Gamma}_i^t(\sigma_t) - \Gamma_i(\sigma_t)\right]\\
&\leq \left[ \Gamma_i(\sigma^*) - \hat{\Gamma}_i^t(\sigma^*) \right] + \left[ \hat{\Gamma}_i^t(\sigma_t) - \Gamma_i(\sigma_t)\right]\\
&\leq \sum_{\sigma\in\lbrace \sigma^*,\sigma_t \rbrace} \left| \hat{\Gamma}_i^t(\sigma) - \Gamma_i(\sigma) \right|\\
&\leq \cfrac{4C_\rho NMK\ln t}{\min_j\mu_{i,j}(1-1/C)t\varepsilon_t},
\end{align*}
where $C_{\rho}\geq 1$ is a constant such that $\|\bm{T}_{i,j}(t)\|_1 = C_{\rho}N_{i,j}(t)$, which depends on the position preference distribution and policy, independent of $K, M, N$.
Then, we evaluate the regret of personalized treatment GreedyRank up to time $t$ as follows
\begin{align*}
&\hspace{13pt}\mathbb{E}[R(t)]\\
&\leq \mathbb{E}[t_0] + \sum_{s=t_0}^t\mathbb{E}\left[ \varepsilon_s + (1-\varepsilon_s)\sum_{i\in[N]}\big(\Gamma_i(\sigma^*) - \Gamma_i(\sigma_s)\big) \right]\\
&\leq \mathbb{E}[t_0] + \sum_{s=t_0}^t\bigg\lbrace\varepsilon_s + \sum_{i\in[N]}\mathbb{E}\left[ \Gamma_i(\sigma^*) - \Gamma_i(\sigma_s) \bigg| \mathcal{F}_s\cap \mathcal{Y}_s^1\cap \ldots\cap\mathcal{Y}_s^N \right]\\
&\hspace{13pt} + \mathbb{P}\left( \mathcal{F}_s^C\right) + \mathbb{P}\left( \mathcal{Y}_s^{1C}\cup\ldots\cup\mathcal{Y}_s^{NC}\right)\bigg\rbrace\\
&\leq \mathbb{E}[t_0] + \sum_{s=t_0}^t\bigg\lbrace\varepsilon_s + \sum_{i\in[N]}\mathbb{E}\left[ \Gamma_i(\sigma^*) - \Gamma_i(\sigma_s) \bigg| \mathcal{F}_s\cap \mathcal{Y}_s^1\cap \ldots\cap\mathcal{Y}_s^N \right]\\
&\hspace{13pt} + \mathbb{P}\left( \mathcal{Y}_s^{1C}\cup\ldots\cup\mathcal{Y}_s^{NC} | \mathcal{A}_s\cap\mathcal{P}_s\cap\mathcal{U}_s\right)+ \mathbb{P}\left( \mathcal{A}_s^C|\mathcal{F}_s\right) + \mathbb{P}\left( \mathcal{P}_s^C|\mathcal{F}_s\right) + \mathbb{P}\left( \mathcal{U}_s^C|\mathcal{F}_s\right) + \mathbb{P}\left( \mathcal{F}_s^C\right)\bigg\rbrace\\
&\leq \mathbb{E}[t_0] \hspace{-2pt}+\hspace{-2pt} \sum_{s=t_0}^t\bigg\lbrace\varepsilon_s \hspace{-2pt}+\hspace{-2pt} \sum_{i\in[N]}\mathbb{E}\left[ \Gamma_i(\sigma^*) \hspace{-2pt}-\hspace{-2pt} \Gamma_i(\sigma_s) \bigg| \mathcal{F}_s\cap \mathcal{Y}_s^1\cap \ldots\cap\mathcal{Y}_s^N \right] \hspace{-2pt}+\hspace{-2pt} N\cdot\mathbb{P}\left( \mathcal{Y}_s^{1C} | \mathcal{A}_s\cap\mathcal{P}_s\cap\mathcal{U}_s\right)\\
&\hspace{12pt}+ \mathbb{P}\left( \mathcal{A}_s^C|\mathcal{F}_s\right) + \mathbb{P}\left( \mathcal{P}_s^C|\mathcal{F}_s\right) + \mathbb{P}\left( \mathcal{U}_s^C|\mathcal{F}_s\cap\mathcal{N}_s \right) + \mathbb{P}\left( \mathcal{N}_s^C\right) + \mathbb{P}\left( \mathcal{F}_s^C\right)\bigg\rbrace\\
&\leq \mathbb{E}[t_0] + \sum_{s=1}^t\bigg\lbrace\varepsilon_s + \cfrac{4C_\rho NMK\ln s}{\min\mu_{i,j}(1-1/C)s\varepsilon_s} + f(1)\delta_s N + N^2Ks^{-2} + s^{-2\ln s} + MKs^{-4}\\
&\hspace{13pt}+ s^{-1/2}\exp(1-s^{-1/2})\ln s + MKs^{-2} + s^{-2} \bigg\rbrace
\end{align*}
Setting $\varepsilon_t=Nt^{-1/2}$, we obtain the regret $\mathbb{E}[R(t)]$ of personalized GreedyRank upper bounded by
\begin{equation*}
\mathbb{E}[R(t)]\leq \sum_{i\in[N]}\sum_{j\in[M]}\cfrac{MK}{\lambda_i\mu_{i,j}} + 2Nt^{1\over 2} + \sum_{i\in[N]}\cfrac{8C_\rho MK\sqrt{\lambda_it}\ln t}{(1-1/C)\min\mu_{i,j}} + \mathcal{O}(1).
\end{equation*}

\end{proof}

\section{Proof of Lemma~\ref{lambda}}
\begin{proof}
At any time $t$, the user $i\in[N]$ arrives with probability $\lambda_i$. Then, by Hoeffding's Inequality, for any $\epsilon\geq 0$, the estimated user arrival rate $\hat{\lambda}_i(t)$ satisfies the following
\begin{equation*}
	\mathbb{P}\left( t\cdot\left| \hat{\lambda}_i(t) - \lambda_i \right|\geq \epsilon \right) \leq \exp\left( -2t\epsilon^2 \right).
\end{equation*}
\end{proof}

\section{Proof of Theorem~\ref{thm_ucb1}}
\RestateUCBa*

\begin{proof}
For user type $i$, define a ``bad'' event $\mathcal{E}^t_i$ regarding policy $\lbrace \sigma_t\rbrace_t$ as follows:
\begin{equation*}
    \mathcal{E}^t_i \triangleq \left\lbrace\hat{\Gamma}_i^t(\sigma_t) + \sum_{j\in\mathrm{M}_{\sigma_t}}\cfrac{a_t\ln t}{N_{i,j}(t)} \geq \hat{\Gamma}^t_i(\sigma^*_i) + \sum_{j\in\mathrm{M}_{\sigma^*_i}}\cfrac{a_t\ln t}{N_{i,j}(t)} \right\rbrace,
\end{equation*}
then event $\mathcal{E}^t_i$ is necessary and sufficient for event $\lbrace \sigma^*_i\neq\sigma_t\rbrace$.
We observe that event $\mathcal{E}^t_i$ implies that at least one of the following events must hold:
\begin{align}
    \hat{\Gamma}^t_i(\sigma^*_t) &\leq \Gamma_i(\sigma^*_i) - \sum_{j\in\mathrm{M}_{\sigma^*_i}}\cfrac{a_t\ln t}{N_{i,j}(t)}\label{1eq1}\\
    \hat{\Gamma}^t_i(\sigma_t) &\geq \Gamma_i(\sigma_t) + \sum_{j\in\mathrm{M}_{\sigma_t}}\cfrac{a_t\ln t}{N_{i,j}(t)}\label{1eq2}\\
    \Gamma(\sigma^*_i) &< \Gamma_i(\sigma_t) + \sum_{j\in\mathrm{M}_{\sigma_t}} \cfrac{2a_t\ln t}{N_{i,j}(t)}\label{1eq3}.
\end{align}

Conditioned on event $\mathcal{Y}_t^1\cap\ldots\cap \mathcal{Y}_t^N$, we obtain that the event (\ref{1eq1}) and (\ref{1eq2}) both happen with probability zero if $a_t\geq 2/\min\mu_{i,j}$.
For event (\ref{1eq3}), when $N_{i,j}(t)= 2K\sqrt{t\ln t}/\min_j\Delta_{i,j}$ and $a_t\leq \sqrt{t/\ln t}$, we have the following with probability one:
\begin{equation*}
    \sum_{j\in\mathrm{M}_{\sigma_t}} \cfrac{2a_t\ln t}{N_{i,j}(t)} =\cfrac{a_t\min_j\Delta_{i,j}}{\sqrt{t/\ln t}} \leq \min_j\Delta_{i,j} \leq \Gamma_i(\sigma^*_i) - \Gamma_i(\sigma_t).
\end{equation*}

Therefore, we obtain the regret of personalized UCBRank $\mathbb{E}[R(t)]$ upper bounded as follows:
\begin{align*}
\mathbb{E}[R(t)]
&= \sum_{s=1}^t\sum_{i\in[N]} \mathbb{E}\left[ \Gamma_i(\sigma^*_i) - \Gamma_i(\sigma_s)\bigg| \mathbb{P}\left( \sigma^*_i\neq \sigma_s \right) \right]\cdot\mathbb{P}\left( \sigma^*_i\neq \sigma_s \right)\nonumber\\
&\leq \mathbb{E}[t_0] + \sum_{s=t_0}^t\bigg( \sum_{i\in[N]}\mathbb{P}\left( \sigma^*_i\neq \sigma_s\big|\mathcal{A}_s\cap\mathcal{P}_s\cap\mathcal{U}_s\cap\mathcal{Y}_s^1\cap \ldots\cap\mathcal{Y}_s^N \right)\\
&\hspace{13pt} + \mathbb{P}(\mathcal{A}^C_s) + \mathbb{P}(\mathcal{P}^C_s) + \mathbb{P}(\mathcal{U}^C_s) + \mathbb{P}\left(\mathcal{Y}^{1C}_s\cup\ldots\cup\mathcal{Y}^{NC}_s\right) \bigg)\\
&\leq \mathbb{E}[t_0] + \sum_{s=t_0}^t\bigg( \sum_{i\in[N]}\mathbb{P}\left( \sigma^*\neq \sigma_s\big|\mathcal{A}_s\cap\mathcal{P}_s\cap\mathcal{U}_s\cap\mathcal{Y}_s^1\cap \ldots\cap\mathcal{Y}_s^N \right)\\
&\hspace{13pt} + \mathbb{P}(\mathcal{A}^C_s) + \mathbb{P}(\mathcal{P}^C_s) + \mathbb{P}(\mathcal{U}^C_s|\mathcal{N}_s) + \mathbb{P}(\mathcal{N}^C_s) + \mathbb{P}\left(\mathcal{Y}^{1C}_s\cup\ldots\cup\mathcal{Y}^{NC}_s\right) \bigg)\\
&\leq \mathbb{E}[t_0] + \sum_{i\in[N]}\cfrac{2C_\rho MK\sqrt{\lambda_it\ln t}}{\min_j\Delta_{i,j}} + \sum_{s=t_1}^t\bigg[ N^2Ks^{-2} + s^{-2\ln s} + MKs^{-4}\\
&\hspace{13pt} + s^{-1/2}\exp(1-s^{-1/2})\ln s + MKs^{-2} \bigg]\\
&\leq \sum_{i\in[N]}\sum_{j\in[M]}\cfrac{MK}{\lambda_i\mu_{i,j}} + \sum_{i\in[N]}\cfrac{2C_\rho MK\sqrt{\lambda_it\ln t}}{\min_j\Delta_{i,j}} + \mathcal{O}(1).
\end{align*}
\end{proof}

\section{Proof of Theorem~\ref{thm_greedy2}}
\RestateGREEDYb*

\begin{proof}
Similar with the proof of Theorem~\ref{thm_greedy1}, we obtain the regret of the initialization phase as upper bounded by
\begin{equation*}
    \mathbb{E}[t_0]\leq \sum_{i\in[N]}\sum_{j\in[M]}\cfrac{MK}{\lambda_i\mu_{i,j}},
\end{equation*}
We now analyze the exploitation.
We first obtain a concentration of the estimated CUF $\hat{\Gamma}_t(\sigma)$.
Since the utility function $f$ is $L_f$-Lipschitz continuous, then for any user $i$ and any policy $\lbrace\sigma_t\rbrace_t$, we have
\begin{align}
    &\hspace{13pt}\mathbb{P}\left( \left|f\left(\hat{\Gamma}_i^t(\sigma_t)\right) \hspace{-2pt}-\hspace{-2pt} f\left(\Gamma_i(\sigma_t)\right)\right| \geq L_f\hspace{-2pt}\sum_{j\in\mathrm{M}_{\sigma_t}}\cfrac{2\ln t}{\mu_{i,j}N_{i,j}(t)} \hspace{4pt}\Bigg|\hspace{4pt} \mathcal{P}_t, \mathcal{U}_t \right)\nonumber\\
    &\leq \mathbb{P}\left( L_f\cdot\left|\hat{\Gamma}_i^t(\sigma_t) - \Gamma_i(\sigma_t)\right| \geq L_f\cdot\sum_{j\in\mathrm{M}_{\sigma_t}}\cfrac{2\ln t}{\mu_{i,j}N_{i,j}(t)} \hspace{4pt}\Bigg|\hspace{4pt} \mathcal{P}_t, \mathcal{U}_t \right)\nonumber\\
    &\leq Kt^{-2}.\label{eqq1}
\end{align}
Define an event $\mathcal{A}_t$ regarding estimated user arrival rate $\hat{\bm{\lambda}}_t$ as follows: at time $t$, for any user $i\in[N]$, there exists $\epsilon_1\geq 0$ such that $|\hat{\lambda}_i(t)-\lambda_i|<\epsilon_1$.
Then, conditioned on event $\mathcal{A}_t\cap\mathcal{P}_t\cap \mathcal{U}_t$, for user $i\in[N]$, we have
\begin{align}
	&\hspace{13pt}\mathbb{P}\left( \hat{\lambda}_i(t)\cdot\left|f\left(\hat{\Gamma}_i^t(\sigma_t)\right) - f\left(\Gamma_i(\sigma_t)\right)\right| \geq \hat{\lambda}_i(t)L_f\cdot\sum_{j\in\mathrm{M}_{\sigma_t}}\cfrac{2\ln t}{\mu_{i,j}N_{i,j}(t)} \hspace{4pt}\Bigg|\hspace{4pt} \mathcal{A}_t, \mathcal{P}_t, \mathcal{U}_t \right)\nonumber\\
	&\geq \mathbb{P}\left( \hat{\lambda}_i(t)\cdot\left|f\left(\hat{\Gamma}_i^t(\sigma_t)\right) - f\left(\Gamma_i(\sigma_t)\right)\right| \geq L_f\cdot\sum_{j\in\mathrm{M}_{\sigma_t}}\cfrac{2\ln t}{\mu_{i,j}N_{i,j}(t)} \hspace{4pt}\Bigg|\hspace{4pt} \mathcal{A}_t, \mathcal{P}_t, \mathcal{U}_t \right)\nonumber\\
	&\geq \mathbb{P}\left( \left|\hat{\lambda}_i(t)f\left(\hat{\Gamma}_i^t(\sigma_t)\right) - (\lambda_i + \epsilon_2)f\left(\Gamma_i(\sigma_t)\right)\right| \geq L_f\sum_{j\in\mathrm{M}_{\sigma_t}}\cfrac{2\ln t}{\mu_{i,j}N_{i,j}(t)} \hspace{4pt}\Bigg|\hspace{4pt} \mathcal{A}_t, \mathcal{P}_t, \mathcal{U}_t \right)\nonumber\\
    &\geq \mathbb{P}\left( \left|\hat{\lambda}_i(t)f\left(\hat{\Gamma}_i^t(\sigma_t)\right) - \lambda_if\left(\Gamma_i(\sigma_t)\right)\right| \geq \epsilon_2f(1) + L_f\sum_{j\in\mathrm{M}_{\sigma_t}}\cfrac{2\ln t}{\mu_{i,j}N_{i,j}(t)} \Bigg| \mathcal{A}_t, \mathcal{P}_t, \mathcal{U}_t \right),\label{eqq2}
\end{align}
and by Eq.~(\ref{eqq1}), we have $(\ref{eqq2}) \leq Kt^{-2}$.
By union bound, we obtain
\begin{equation}
\label{gamma_concentration_org}
    \mathbb{P}\left( \left|\hat{\Gamma}_t(\sigma_t) - \Gamma(\sigma_t)\right|\geq \sum_{i\in[N]}\epsilon_1f(1) + L_f\sum_{i\in[N]}\sum_{j\in\mathrm{M}_{\sigma_t}}\cfrac{2\ln t}{\mu_{i,j}N_{i,j}(t)} \hspace{4pt}\bigg|\hspace{4pt} \mathcal{A}_t, \mathcal{P}_t, \mathcal{U}_t\right) \leq NKt^{-2}.
\end{equation}
Specifically, setting $\epsilon_1=\sum_{j\in\mathrm{M}_{\sigma_t}}2L_f\ln t/\left(f(1)\mu_{i,j}N_{i,j}(t)\right)$, we obtain
\begin{equation}
\label{gamma_concentration}
	\mathbb{P}\left( \left|\hat{\Gamma}_t(\sigma_t) - \Gamma(\sigma_t)\right|\geq 2L_f\sum_{i\in[N]}\sum_{j\in\mathrm{M}_{\sigma_t}}\cfrac{2\ln t}{\mu_{i,j}N_{i,j}(t)} \hspace{4pt}\bigg|\hspace{4pt} \mathcal{A}_t, \mathcal{P}_t, \mathcal{U}_t\right) \leq NKt^{-2},
\end{equation}
in which case, we have the probability of event $\mathbb{P}(\mathcal{A}_t^C)\leq t^{-2C_1\ln t}$ with constant $C_1>0$.

We proved that in policy GreedyRank, conditioned on event $\mathcal{F}_t$, for user $i$ and arm $j$, we have
\begin{equation*}
	\mathbb{P}\left( t\varepsilon_t - \xi_t \geq \cfrac{t\varepsilon_t}{C}\right)\leq \exp\left(-\cfrac{2t\varepsilon_t^2}{C^2}\right)\leq \exp\left( -2t^{1/2} \right).
\end{equation*}
Then, define an event $\mathcal{Y}_t$ regarding estimated CUF $\hat{\Gamma}_t(\sigma)$ as follows: for any policy $\lbrace \sigma_t\rbrace_t$, it holds that $|\hat{\Gamma}_t(\sigma) - \Gamma(\sigma_t)| < 2L_f\sum_{i\in[N]}\sum_{j\in\mathrm{M}_{\sigma_t}}2\ln t/(\mu_{i,j}N_{i,j}(t))$.
Conditioned on event $\mathcal{F}_t\cap \mathcal{Y}_t$, we have
\begin{equation*}
\Gamma(\sigma^*) - \Gamma(\sigma_t) = \left[ \Gamma(\sigma^*) - \hat{\Gamma}_t(\sigma^*) \right] + \left[ \hat{\Gamma}_t(\sigma^*) - \hat{\Gamma}_t(\sigma_t)\right] + \left[ \hat{\Gamma}_t(\sigma_t) - \Gamma(\sigma_t)\right].
\end{equation*}

Specifically, $\hat{\Gamma}_t(\sigma_t)$ is defined as being maximized by permutation $\sigma_t$. If we use an approximate solution with a factor $\delta_t$ of being optimal, then we have
\begin{align*}
\Gamma(\sigma^*) - \Gamma(\sigma_t)
&\leq \left[ \Gamma(\sigma^*) - \hat{\Gamma}_t(\sigma^*) \right] + \left[ \hat{\Gamma}_t(\sigma^*) - (1-\delta_t)\hat{\Gamma}_t(\sigma_t)\right] + \left[ \hat{\Gamma}_t(\sigma_t) - \Gamma(\sigma_t)\right]\\
&\leq f(1)\delta_t N + \sum_{\sigma\in\lbrace \sigma^*, \sigma_t \rbrace} \left| \Gamma(\sigma) - \hat{\Gamma}_t(\sigma) \right|\\
&\leq f(1)\delta_t N + \cfrac{4L_fC_\rho N^2MK\ln t}{\min\mu_{i,j}(1-1/C)t\varepsilon_t}.
\end{align*}
Then, we evaluate the regret of equal treatment GreedyRank up to time $t$ as follows
\begin{align*}
\mathbb{E}[R(t)]
&\leq \mathbb{E}[t_0] + \sum_{s=t_0}^t\mathbb{E}\left[ \varepsilon_s + (1-\varepsilon_s)\big(\Gamma(\sigma^*) - \Gamma(\sigma_s)\big) \right]\\
&\leq \mathbb{E}[t_0] + \sum_{s=t_0}^t\bigg\lbrace\varepsilon_s + \mathbb{E}\left[ \Gamma(\sigma^*) - \Gamma(\sigma_s) \bigg| \mathcal{F}_s\cap \mathcal{Y}_s \right] + \mathbb{P}\left( \mathcal{F}_s^C\right) + \mathbb{P}\left( \mathcal{Y}_s^C\right)\bigg\rbrace\\
&\leq \mathbb{E}[t_0] + \sum_{s=t_0}^t\bigg\lbrace\varepsilon_s + \mathbb{E}\left[ \Gamma(\sigma^*) - \Gamma(\sigma_s) \bigg| \mathcal{F}_s\cap \mathcal{Y}_s \right] + \mathbb{P}\left( \mathcal{Y}_s^C | \mathcal{A}_s\cap\mathcal{P}_s\cap\mathcal{U}_s\right)\\
&\hspace{12pt}+ \mathbb{P}\left( \mathcal{A}_s^C|\mathcal{F}_s\right) + \mathbb{P}\left( \mathcal{P}_s^C|\mathcal{F}_s\right) + \mathbb{P}\left( \mathcal{U}_s^C|\mathcal{F}_s\right) + \mathbb{P}\left( \mathcal{F}_s^C\right)\bigg\rbrace\\
&\leq \mathbb{E}[t_0] + \sum_{s=t_0}^t\bigg\lbrace\varepsilon_s + \mathbb{E}\left[ \Gamma(\sigma^*) - \Gamma(\sigma_s) \bigg| \mathcal{F}_s\cap \mathcal{Y}_s \right] + \mathbb{P}\left( \mathcal{Y}_s^C | \mathcal{A}_s\cap\mathcal{P}_s\cap\mathcal{U}_s\right)\\
&\hspace{12pt}+ \mathbb{P}\left( \mathcal{A}_s^C|\mathcal{F}_s\right) + \mathbb{P}\left( \mathcal{P}_s^C|\mathcal{F}_s\right) + \mathbb{P}\left( \mathcal{U}_s^C|\mathcal{F}_s\cap\mathcal{N}_s \right) + \mathbb{P}\left( \mathcal{N}_s^C\right) + \mathbb{P}\left( \mathcal{F}_s^C\right)\bigg\rbrace\\
&\leq \mathbb{E}[t_0] + \sum_{s=1}^t\bigg\lbrace\varepsilon_s + \cfrac{4L_fC_\rho N^2MK\ln s}{\min\mu_{i,j}(1-1/C)s\varepsilon_s} + f(1)\delta_s N + NKs^{-2} + s^{-2\ln s}\\
&\hspace{13pt}+ MKs^{-4}+ s^{-1/2}\exp(1-s^{-1/2})\ln s + MKs^{-2} + s^{-2} \bigg\rbrace
\end{align*}
Setting $\varepsilon_t=\Theta(N\cdot t^{-1/2})$, we obtain the regret $\mathbb{E}[R(t)]$ of policy GreedyRank upper bounded by
\begin{equation*}
\mathbb{E}[R(t)]\leq \sum_{i\in[N]}\sum_{j\in[M]}\cfrac{MK}{\lambda_i\mu_{i,j}} + 2Nt^{1\over 2} + \cfrac{8L_fC_\rho NMK}{(1-1/C)\min\mu_{i,j}}t^{1\over 2}\ln t + f(1)\delta_t Nt + \mathcal{O}(1).
\end{equation*}
\end{proof}

\section{Proof of Theorem~\ref{thm_ucb2}}

\RestateUCBb*

\begin{proof}
Define a ``bad'' event $\mathcal{E}_t$ regarding policy $\lbrace \sigma_t\rbrace_t$ as follows:
\begin{equation*}
	\mathcal{E}_t \triangleq \left\lbrace\hat{\Gamma}_t(\sigma_t) + \sum_{i\in[N]}\sum_{j\in\mathrm{M}_{\sigma_t}}\cfrac{a_t\ln t}{N_{i,j}(t)} \geq \hat{\Gamma}_t(\sigma^*) + \sum_{i\in[N]}\sum_{j\in\mathrm{M}_{\sigma^*}}\cfrac{a_t\ln t}{N_{i,j}(t)} \right\rbrace,
\end{equation*}
then event $\mathcal{E}_t$ is necessary and sufficient for event $\lbrace \sigma^*\neq\sigma_t\rbrace$.
We observe that event $\mathcal{E}_t$ implies that at least one of the following events must hold:
\begin{align}
	\hat{\Gamma}_t(\sigma^*) &\leq \Gamma(\sigma^*) - \sum_{i\in[N]}\sum_{j\in\mathrm{M}_{\sigma^*}}\cfrac{a_t\ln t}{N_{i,j}(t)}\label{eq1}\\
	\hat{\Gamma}_t(\sigma_t) &\geq \Gamma(\sigma_t) + \sum_{i\in[N]}\sum_{j\in\mathrm{M}_{\sigma_t}}\cfrac{a_t\ln t}{N_{i,j}(t)}\label{eq2}\\
	\Gamma(\sigma^*) &< \Gamma(\sigma_t) + \sum_{i\in[N]}\sum_{j\in\mathrm{M}_{\sigma_t}} \cfrac{2a_t\ln t}{N_{i,j}(t)}\label{eq3}.
\end{align}

Conditioned on event $\mathcal{Y}_t$, we obtain that the event (\ref{eq1}) and (\ref{eq2}) both happen with probability zero if $a_t\geq 4L_f/\min\mu_{i,j}$ and the following condition is satisfied:
\begin{equation}
\label{delta}
	\delta_t \leq \cfrac{a_t-4L_f/\min\mu_{i,j}}{f(1)N}\cdot\sum_{i\in[N]}\sum_{j\in\mathrm{M}_{\sigma_t}}\cfrac{\ln t}{N_{i,j}(t)}.
\end{equation}

For event (\ref{eq3}), when $N_{i,j}(t)= 2NK\sqrt{t\ln t}/\Delta_{\Gamma}$ and $a_t\leq \sqrt{t/\ln t}$, we have the following with probability one:
\begin{equation*}
	\sum_{i\in[N]}\sum_{j\in\mathrm{M}_{\sigma_t}} \cfrac{2a_t\ln t}{N_{i,j}(t)} =\cfrac{a_t\Delta_{\Gamma}}{\sqrt{t/\ln t}} \leq \Delta_{\Gamma} \leq \Gamma(\sigma^*) - \Gamma(\sigma_t).
\end{equation*}

Therefore, if $\|\bm{T}_{i,j}(t)\|_1\geq 2C_\rho NK\sqrt{t\ln t}/\Delta_{\Gamma}$, $a_t\in(2L_f/\min\mu_{i,j}, \sqrt{t/\ln t}]$ and condition (\ref{delta}) is satisfied, we obtain the regret of equal treatment UCBRank $\mathbb{E}[R(t)]$ upper bounded as follows:
\begin{align*}
\mathbb{E}[R(t)]
&= \sum_{s=1}^t \mathbb{E}\left[ \Gamma(\sigma^*) - \Gamma(\sigma_s)\bigg| \mathbb{P}\left( \sigma^*\neq \sigma_s \right) \right]\cdot\mathbb{P}\left( \sigma^*\neq \sigma_s \right)\nonumber\\
&\leq \mathbb{E}[t_0] + \sum_{s=t_0}^t\bigg\lbrace f(1)N\cdot \bigg( \mathbb{P}\left( \sigma^*\neq \sigma_s\big|\mathcal{A}_s\cap\mathcal{P}_s\cap\mathcal{U}_s\cap\mathcal{Y}_s \right)\\
&\hspace{13pt} + \mathbb{P}(\mathcal{A}^C_s) + \mathbb{P}(\mathcal{P}^C_s) + \mathbb{P}(\mathcal{U}^C_s) + \mathbb{P}(\mathcal{Y}^C_s) \bigg) \bigg\rbrace\\
&\leq \mathbb{E}[t_0] + \sum_{s=t_0}^t\bigg\lbrace f(1)N\cdot \bigg( \mathbb{P}\left( \sigma^*\neq \sigma_s\big|\mathcal{A}_s\cap\mathcal{P}_s\cap\mathcal{U}_s \right)\\
&\hspace{13pt} + \mathbb{P}(\mathcal{A}^C_s) + \mathbb{P}(\mathcal{P}^C_s) + \mathbb{P}(\mathcal{U}^C_s|\mathcal{N}_s) + \mathbb{P}(\mathcal{N}^C_s) + \mathbb{P}(\mathcal{Y}^C_s) \bigg) \bigg\rbrace\\
&\leq \mathbb{E}[t_0] + \cfrac{2C_\rho N^2MK\sqrt{t\ln t}}{\Delta_{\Gamma}} + \sum_{s=t_1}^t\bigg\lbrace f(1)N \bigg( NKs^{-2} + s^{-2\ln s}\\
&\hspace{13pt} + MKs^{-4} + s^{-1/2}\exp(1-s^{-1/2})\ln s + MKs^{-2} \bigg) \bigg\rbrace\\
&\leq \sum_{i\in[N]}\sum_{j\in[M]}\cfrac{MK}{\lambda_i\mu_{i,j}} + \cfrac{2C_\rho N^2MK\sqrt{t\ln t}}{\Delta_{\Gamma}} + C_1f(1)N\ln t +  \mathcal{O}(1).
\end{align*}
\end{proof}

\end{document}